\documentclass{article}

\usepackage[final]{corl_2026} 
\usepackage{booktabs}
\usepackage{graphicx}
\usepackage{amssymb}
\usepackage{amsmath}
\usepackage[normalem]{ulem}
\usepackage{placeins}
\usepackage{algorithm}
\usepackage{algpseudocode}
\usepackage{multirow}
\usepackage{enumitem}
\usepackage{dblfloatfix}

\title{GLST: Global-Local Spatio-Temporal \\ Memory Decomposition Framework for Online \\ Chained-Goal Navigation}

\author{
{\bfseries
\begin{tabular}{ccccc}
Yousef Liang$^{1}$ &
Ryan Ardalan$^{2}$ &
Yarden Goraly$^{2}$ &
Chenhao Zhu$^{1}$ &
Yue Yin$^{1}$
\end{tabular}
}
\\[0.5ex]
{\bfseries
\begin{tabular}{cccc}
Jonas Frey$^{1,2}$ &
Claire Tomlin$^{2}$ &
Marco Pavone$^{1,3}$ &
Shankar Sastry$^{2}$
\end{tabular}
}
\\[2ex]
$^{1}$Stanford University \qquad
$^{2}$University of California, Berkeley \qquad
$^{3}$NVIDIA
\\[0.5ex]
\texttt{youzhil@stanford.edu}
}

\begin{document}
\maketitle


\begin{abstract}
Autonomous navigation in real-world environments often requires long-horizon operation under partial observability and evolving goals. Existing mapless navigation approaches predominantly assume fixed goals throughout an episode, limiting applicability to practical scenarios where goals change online during execution. We study online chained-goal navigation, where goals evolve through existing-goal redirection or new-goal emergence without explicit transition signals.
Under this setting, we find navigation may require either continued exploration or strategic revisitation, creating conflicting memory requirements. We propose Global-Local Spatio-Temporal Belief Learning (GLST), a memory-decomposition framework that separates latent beliefs into global-local spatial and long-short temporal representations for long-horizon reasoning. We further introduce goal-conditioned rewards to balance exploration and exploitation under evolving goals. Extensive experiments in simulation across diverse terrains demonstrate improved chained-goal navigation performance over baselines. Real-robot experiments using the Spot quadruped further validate our memory decomposition framework. 

\end{abstract}

\keywords{Online Chained-Goal Navigation, Spatio-Temporal Memory Decomposition, Reinforcement Learning} 


\section{Introduction}


Autonomous navigation in real-world environments often requires operating under partial observability, long horizons, and evolving objectives~\cite{fan2026long, yang2025spatially, mirowski2016learning}. 
In practical deployments such as search-and-rescue and emergency response scenarios, navigation goals may change online due to shifting target priorities, updated target localizations, or newly discovered target identities~\cite{feng2025hierarchical, ghassemi2022multi}.
Consequently, unlike conventional navigation settings with fixed destinations, the navigation objective itself becomes temporally evolving throughout execution.

Despite substantial progress in mapless navigation, existing reinforcement learning and memory-based frameworks predominantly assume a fixed navigation objective throughout an episode~\cite{zhou2022deep, schmittle2025long, zhang2025dimension}. While such formulations are useful for evaluating navigation and locomotion capabilities, they overlook an important class of real-world problems where goals evolve online.

In this work, we study online chained-goal navigation, where navigation objectives evolve during execution. We consider two forms of goal transitions: (i) \emph{existing-goal redirection}, where a new goal is issued before the current goal is completed, and (ii) \emph{new-goal emergence}, where a new goal is assigned after completion of the current goal. The former often corresponds to target updates or relocation, while the latter arises from assignment of new objectives such as target discovery or emergency events. The agent may not be explicitly informed which transition mechanism applies and must handle both cases using only historical observations and interactions; we refer to this setting as navigation \emph{without explicit transition signals}.

These challenges expose limitations of existing navigation paradigms. Restarting navigation after each goal update leads to compounding failures across chained objectives and exponential degradation in long-horizon success rates (Appendix~\ref{app:chained_sr}). Alternatively, maintaining a unified memory may induce conflicts between exploration behaviors that encourage discovering unseen regions and revisitation behaviors that strategically exploit overlaps with previously explored regions. 

We propose a memory-decomposition framework 
under evolving goals. Our approach decomposes memory into global-local spatial and long-short temporal representations, preserving long-horizon context while supporting local adaptation (\href{https://github.com/Yousef-Liang/glst}{Code Repo}). Specifically, our contributions include:
\begin{enumerate}[label=(\roman*)]
    \item We formulate online chained-goal navigation as an end-to-end partially observable navigation problem with ambiguous goal evolution during execution.
    \item We identify conflicting exploration-revisitation objectives and propose a goal-conditioned global-local spatial and long-short temporal decomposition with tailored rewards.    
    \item We validate the proposed framework through extensive simulations across diverse terrains and horizons, demonstrating robust navigation under evolving goals, and further validate it through real-robot deployments.

\end{enumerate}

\section{Related Work}

\textbf{Long-Range Mapless Navigation} Mapless navigation has been extensively studied using learning-based approaches, particularly reinforcement learning~\cite{cai2024deep, zheng2025embodied} and imitation learning~\cite{naveed2025help, kim2024visual}, where end-to-end visuomotor policies map onboard observations to control actions without requiring explicit metric maps~\cite{kok2026reinforcement, mohamed2026wheeled}. To address long-horizon navigation under partial observability, recent approaches incorporate memory into navigation systems or policies to capture historical context and maintain latent representations of past observations~\cite{zhang2026bmtd3, yang2023spiking, han2025real, johnson2024memory}. These methods improve robustness and exploration under partial observability~\cite{yang20252, liu2024cognitive}. However, existing formulations predominantly assume a single static goal throughout an episode, under which memory primarily supports state inference and exploration rather than strategic revisitation or reasoning under evolving navigation objectives.

\textbf{Dynamic Goal Navigation} Dynamic-goal navigation studies navigation toward persistent targets with changing states or locations over time~\cite{feng2025hierarchical, li2023predictive}. Existing approaches commonly focus on target localization and target following under uncertainty~\cite{xu2011target,shurrab2023reinforcement, masmitja2023dynamic}. Recent advances leverage trajectory prediction methods ranging from dynamics-based estimation and filtering approaches~\cite{elnagar2001prediction, wu2010moving, abdelkader2024smart} to learning-based prediction models~\cite{liu2025vehicle,lu2025hyper, li2024multi} for improving localization, planning, and target-following performance. While related, these formulations assume persistent targets with predictable evolution, whereas our setting allows goals to change discontinuously across arbitrary locations.


\textbf{Multiple Goal Navigation} Multiple-goal navigation studies tasks involving multiple objectives or intermediate targets. Existing approaches commonly formulate navigation as visiting a set of goals or waypoints, where the primary challenge lies in determining efficient goal ordering, route planning, or hierarchical subgoal generation~\cite{lee2024learning, gorchetchnikov2005biophysical, rao2021visual, gao2025hierarchical}. Three typical scenarios involve map-based optimization of goal ordering and route planning~\cite{gorchetchnikov2005biophysical}, multiple-goal navigation through independent goal executions~\cite{rao2021visual}, and waypoint sequencing~\cite{kivrak2022waypoint}. However, these formulations typically assume predefined goal structures and explicit subgoal transitions, with limited emphasis on memory representations for long-horizon exploration and efficient revisitation across evolving objectives.

\textbf{Research Gap} 
Existing methods independently address memory, dynamic targets, or multi-goal planning, but do not study how evolving objectives alter the role of navigation memory itself.
In particular, prior methods do not explicitly address how evolving objectives interact with navigation memory, where exploration-oriented behaviors may conflict with strategic revisitation across chained goals. 

\section{Methods}

\subsection{Problem Formulation}

We formulate online chained-goal navigation for a quadruped robot in 3D environments as a partially observable Markov decision process (POMDP), defined by hidden states $s_t$, observations $o_t$, actions $a_t$, transition dynamics, and rewards $\mathcal{R}$. During navigation, both online existing-goal redirection and new-goal emergence may occur without explicit transition signals. Consequently, the policy must handle both cases using only observation histories of time-varying goals, environment structure, and robot states. We model the goal $g_{n, t}$ as a time-varying observation at timestep $t$ for the $n$-th goal.

To improve tractability over long horizons, we adopt a hierarchical formulation. A high-level navigation policy $\pi_{\mathrm{nav}}$ outputs body-frame velocity commands $a_t=(v_x,v_y,\omega_z)$. 
These commands are executed by a low-level locomotion controller $\pi_{\mathrm{loc}}$, which maps navigation commands, exteroceptive observations, and proprioceptive observations into joint actions. 

The objective is to learn a navigation policy that maximizes the expected cumulative return over chained goals during execution
:
\begin{align*}
\pi_{\mathrm{nav}}^{*}
&=
\arg\max_{\pi} J(\theta)
=
\arg\max_{\pi}
\underset{\tau}{\mathbb{E}}
\left[
\sum_{n=1}^{N_g}
\sum_{t=1}^{T_n}
\mathcal{R}_{n,t}
\right],
\end{align*}
where $J(\theta)$ denotes the objective parameterized by $\theta$, and
$\tau=\{\tau_n\}_{n=1}^{N_g}$ denotes the full chained trajectory. Each segment
$\tau_n=\{x_{n,t}\}_{t=1}^{T_n}$ is associated with the $n$-th goal, where
$x_{n,t}\in\mathbb{R}^3$ is the robot position and $T_n$ is the segment horizon.


To escape local optima, navigation methods commonly introduce exploration rewards that discourage revisiting previously explored states. 
The collected rewards depend on trajectories induced by policies conditioned on latent belief representations, where $\mathcal{E}$ encodes observation histories into latent states:
$
\mathcal{R}^{+}_{n,t}
=
\mathcal{R}^{+}_{n,t}
\bigl(
o_{n,t}, 
\pi_\theta( \mathcal{E}(o_{1:n,1:t-1}) )
\bigr).
$
In chained-goal navigation, however, trajectory overlap naturally arises. Efficient navigation may therefore require revisitation: 
$
\mathcal{R}^{-}_{n,t}
=
\mathcal{R}^{-}_{n,t}
\bigl(
o_{n,t}, 
\pi_\theta( \mathcal{E}(o_{1:n,1:t-1}) )
\bigr).
$

A trajectory-level exploration objective can be expressed in the following general form:
\begin{align*}
\mathcal{R}^{+}(\tau_n)
&=
\lambda_{+}
\sum_{t=1}^{T_n}
\sum_{(m,k)\prec(n,t)}
D_{\mathrm{JS}}
\left(
p_{n,t}
\|
p_{m,k}
\right),
\end{align*}
where $\lambda_+>0$ is the exploration weight, $p_{n,t}(x)=\kappa(x,x_{n,t})$ denotes the kernel distribution associated with position $x_{n,t}$, $(m,k)\prec(n,t)$ denotes lexicographic precedence, and $D_{\mathrm{JS}}(\cdot\|\cdot)$ is the Jensen-Shannon divergence. Similarly, the revisitation objective can be expressed as
\begin{align*}
\mathcal{R}^{-}(\tau_n)
&=
-\lambda_{-}
\sum_{t=1}^{T_n}
\min_{1\le j\le L_n}
D_{\mathrm{JS}}
\left(
p_{n,t}
\|
p^{*}_{n,j}
\right),
\end{align*}
where $\lambda_->0$ is the revisitation weight and
$
p^{*}_{n,j}(x)=\kappa(x,x^{*}_{n,j})
$
denotes the kernel distribution centered at the $j$-th state of the optimal reference trajectory.

In chained-goal navigation, goals may change arbitrarily during execution, requiring either continued exploration or revisitation depending on historical context and spatial relationships with previously explored regions. 
When optimal trajectories overlap with previously visited regions, exploration discourages revisitation while exploitation encourages it, inducing opposing local gradient directions:
\begin{align*}
\left\langle
\nabla \mathcal{J}^{+},
\nabla \mathcal{J}^{-}
\right\rangle
<
0.
\end{align*}
A detailed derivation of the resulting gradient conflict is provided in Appendix~\ref{app:conflicting_gradients}.

\begin{figure}[t]
    \centering
    \includegraphics[width=1\linewidth]{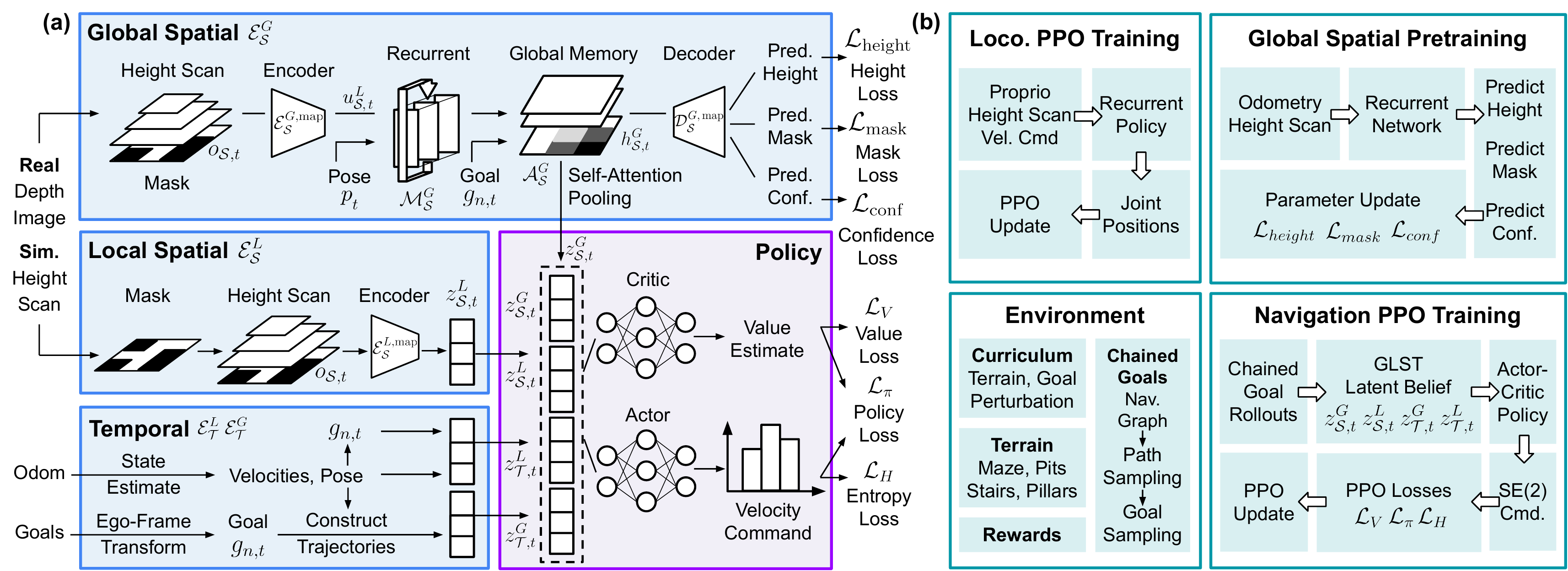}
\caption{
Overall framework and training pipeline of the proposed GLST method. (a) GLST framework showing the global-local spatial and temporal belief branches and actor-critic policy. Real-robot depth images or simulated height scans serve as shared inputs to both spatial branches. (b) Training pipeline including locomotion policy training, global spatial supervised pretraining, environment generation with chained-goal sampling, and navigation policy training.
}
    \label{fig:architecture_compact}
\end{figure}

This gradient conflict makes joint learning inefficient under a shared memory representation. To mitigate this issue, we introduce separated goal-conditioned memory encoders $\mathcal{E}^{+}_{n,t}$, $\mathcal{E}^{-}_{n,t}$ and rewards $\mathcal{R}^{+}_{n,t}$, $\mathcal{R}^{-}_{n,t}$ for exploration and revisitation:
\begin{equation*}
\begin{array}{r@{\;}c@{\;}l@{\qquad\qquad}r@{\;}c@{\;}l}
\mathcal{E}^{+}_{n,t}
&=&
\mathcal{E}^{+}\left(o_{1:n,1:t-1}\mid g_{n,t}\right),
&
\mathcal{E}^{-}_{n,t}
&=&
\mathcal{E}^{-}\left(o_{1:n,1:t-1}\mid g_{n,t}\right);
\\[0.5em]
\mathcal{R}^{+}_{n,t}
&=&
\mathcal{R}^{+}_{n,t}\left(o_{n,t}, \pi_\theta(\mathcal{E}^{+}_{n,t}) \mid g_{n,t} \right),
&
\mathcal{R}^{-}_{n,t}
&=&
\mathcal{R}^{-}_{n,t}\left(o_{n,t}, \pi_\theta(\mathcal{E}^{-}_{n,t}) \mid g_{n,t} \right).
\end{array}
\end{equation*}

Shared latent representations introduce information mixing and loss of trajectory-specific context. Rather than using a unified encoder over all histories, we design goal-conditioned subtrajectory-aware encoders to separate signals associated with different trajectory segments. Unlike RL methods with unified memory representations that balance competing objectives, exploration and revisitation may require different memory structures depending on context, making compromised representations suboptimal. Instead, we explicitly decompose memory representations to preserve context-dependent information and utilization.

\FloatBarrier
\subsection{Global-Local Spatio-Temporal Belief Learning (GLST)}

The goal-conditioned GLST framework, shown in Fig.~\ref{fig:architecture_compact}(a), decomposes latent beliefs into global-local spatial and long-short temporal representations, enabling joint reasoning over local traversability, global environment structure, short-horizon dynamics, and long-horizon revisitation. 

\textbf{Global and Local Spatial Encoder} The spatial module consists of local and global spatial belief encoders, $\mathcal{E}_{\mathcal{S}}^L$ and $\mathcal{E}_{\mathcal{S}}^G$, respectively. Local terrain observations are represented as a local height map with a binary visibility mask, constructed from depth observations or simulated height scans. The local spatial belief, $z_{\mathcal{S},t}^L = \mathcal{E}_{\mathcal{S}}^{L,\mathrm{map}}(o_{\mathcal{S},t})$, encodes instantaneous terrain observations around the robot at fine spatial granularity. This representation captures nearby traversability 
for reactive locomotion and local motion planning. A separate local spatial latent representation, $u_{\mathcal{S},t}^L = \mathcal{E}_{\mathcal{S}}^{G,\mathrm{map}}(o_{\mathcal{S},t})$, is accumulated into a pose-conditioned latent global map with a recurrent network, since local and global reasoning require different spatial granularity and memory retention properties. The global spatial memory is recursively updated according to 
$h^{G}_{\mathcal{S},t} = \mathcal{M}_{\mathcal{S}}^G \left( h^{G}_{\mathcal{S},t-1}, u^{L}_{\mathcal{S},t}, p_t \right), $ 
where $p_t$ denotes the robot pose at time step $t$. The accumulated latent global map is then processed by a goal-conditioned self-attention encoder to produce the global spatial belief representation with attention pooling: 
$ z^{G}_{\mathcal{S},t} = \mathcal{A}_{\mathcal{S}}^G \left( h^{G}_{\mathcal{S},t}, g_{n, t} \right). $ 
To preserve meaningful spatial structure, the global spatial branch additionally includes auxiliary prediction heads for map reconstruction, visibility mask, and confidence estimation. 

\textbf{Global and Local Temporal Encoder.}
The temporal module consists of local and global temporal belief encoders,
$\mathcal{E}_{\mathcal{T}}^L$ and $\mathcal{E}_{\mathcal{T}}^G$,
respectively. The local temporal belief,
$z_{\mathcal{T},t}^L$,
captures short-horizon motion dynamics from instantaneous robot states, including velocities and goal observations. In contrast,
$z_{\mathcal{T},t}^G$,
encodes long-horizon trajectory histories across chained goals. Trajectory memories are constructed from historical robot positions and goal observations, while maintaining short-horizon buffers for the current goal and persistent buffers across goal transitions.

The resulting global-local spatio-temporal belief representation is
$
z_t
=
\{
z^{L}_{\mathcal{S},t},
z^{G}_{\mathcal{S},t},
z^{L}_{\mathcal{T},t},
z^{G}_{\mathcal{T},t}
\},
$
which is processed by a shared actor-critic network for action generation and value estimation. 
The overall training pipeline is illustrated in Fig.~\ref{fig:architecture_compact}(b). The framework adopts a staged training procedure in which the locomotion policy and global spatial module are trained separately before navigation training, allowing pretrained low-level control and latent spatial representations to support subsequent policy learning. Architecture, implementation, and goal-sampling details, including configurable constraints, are provided in Appendix~\ref{app:architecture}.

\subsection{Rewards and Curriculum}

In online chained-goal navigation, rewards should provide dense feedback while remaining consistent across evolving goals. The reward function consists of goal-progress rewards, exploration rewards, gait stability rewards, task completion rewards, and collision and excessive-roll penalties. Here we focus on goal-progress and exploration rewards, while complete formulations and curriculum details are provided in Appendix~\ref{app:rewards_and_curriculum}.

To provide dense, directionally consistent feedback, we introduce a goal-progress reward based on the closest achieved distance to the current goal. Privileged segment information is used only for reward construction. Let the robot-relative goal distance be
$
d_{n,t}^{xyz}=\|g_{n,t}\|_2,
$
and define the running minimum distance within segment $n$ as
$
d_{\min,n,t}^{xyz}
=
\min_{\tau\le t}
d_{n,\tau}^{xyz}.
$
The progress reward is
\begin{align*}
\mathcal{R}_{p}^{xyz}(t)
=
\lambda_{p}
\max
\left(
d_{\min,n,t-1}^{xyz}
-
d_{n,t}^{xyz},
0
\right).
\end{align*}
This reward reinforces only newly achieved closest distances, encouraging monotonic progress while reducing oscillation and regressions. To account for differing horizontal and vertical scales, we additionally define
$
\mathcal{R}_{p}^{z}(t)
=
\lambda_{p}^{z}
\max(d_{\min,n,t-1}^{z}-d_{n,t}^{z},0).
$

To discourage stalling and encourage exploration, we introduce a goal-conditioned visitation penalty using a sparse visitation buffer $\mathcal B_n$ maintained for each goal segment. Let $n_{n,t}$ denote the number of nearby visited states around current position $\mathbf x_t$. The exploration penalty can be expressed in a generalized continuous form as
\begin{align*}
\mathcal{R}_e(t)
=
-
\lambda_e
\left(
\min(n_{n,t},n_{\max})
\right)^{\gamma}
\sum_{\mathbf{x}_i\in\mathcal B_n}
\mathcal{N}
\left(
\|\mathbf{x}_t-\mathbf{x}_i\|_2;
0,
\sigma_e^2
\right),
\end{align*}
where $\lambda_e$, $\gamma$, and $\sigma_e$ control penalty magnitude, revisit scaling, and spatial decay. The formulation and its parameters can be adapted to different terrains and robot behaviors.

\section{Experiments}

To validate our memory architecture motivated by conflicting memory gradients, we evaluate our approach in simulation and real-world environments. Simulation experiments are conducted in NVIDIA Isaac Lab~\cite{mittal2025isaac}, while real-world deployment is evaluated using the Spot quadruped in indoor and outdoor environments.

\begin{figure}[t]
    \centering
    \includegraphics[width=1\linewidth]{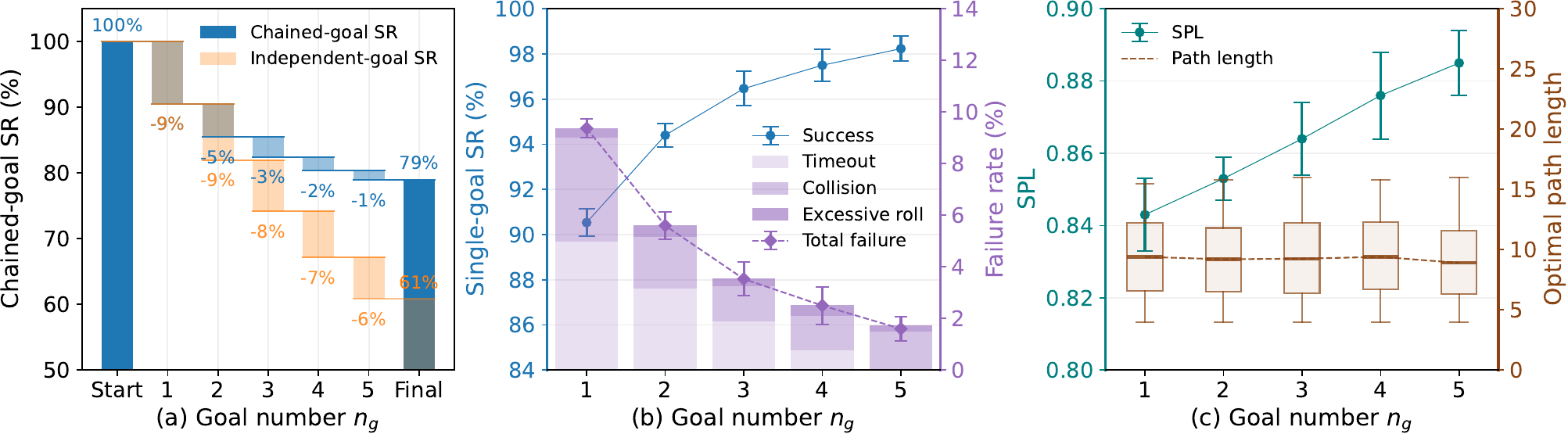}
\caption{
Performance on chained-goal navigation. (a) Chained-goal success rate (SR), where empirical compounding is compared with independent-goal assumptions. (b) Single-goal SR and failure decomposition into timeout, collision, and excessive roll. (c) Navigation efficiency measured by SPL and optimal path length distributions. Increasing SPL across goals suggests improved efficiency under comparable task difficulty.
}
    \label{fig:ours_performance_breakdown}
\end{figure}

\subsection{Simulation Experiments}

We evaluate navigation performance using success rate (SR) and success weighted by path length (SPL). SR is defined as
$
\mathrm{SR}(\%)=
\frac{1}{N}
\sum_{i=1}^{N}
S_i\times100,
$
where $S_i\in\{0,1\}$ indicates successful completion and $N$ is the number of evaluation episodes. SPL measures path efficiency,
$
\mathrm{SPL}
=
\frac{1}{N}
\sum_{i=1}^{N}
S_i
\frac{l_i}{\max(p_i,l_i)},
$
where $l_i$ and $p_i$ denote shortest and traversed path lengths, respectively~\cite{anderson2018evaluation}.

\begin{figure}[t]
    \centering
    \includegraphics[width=\linewidth]{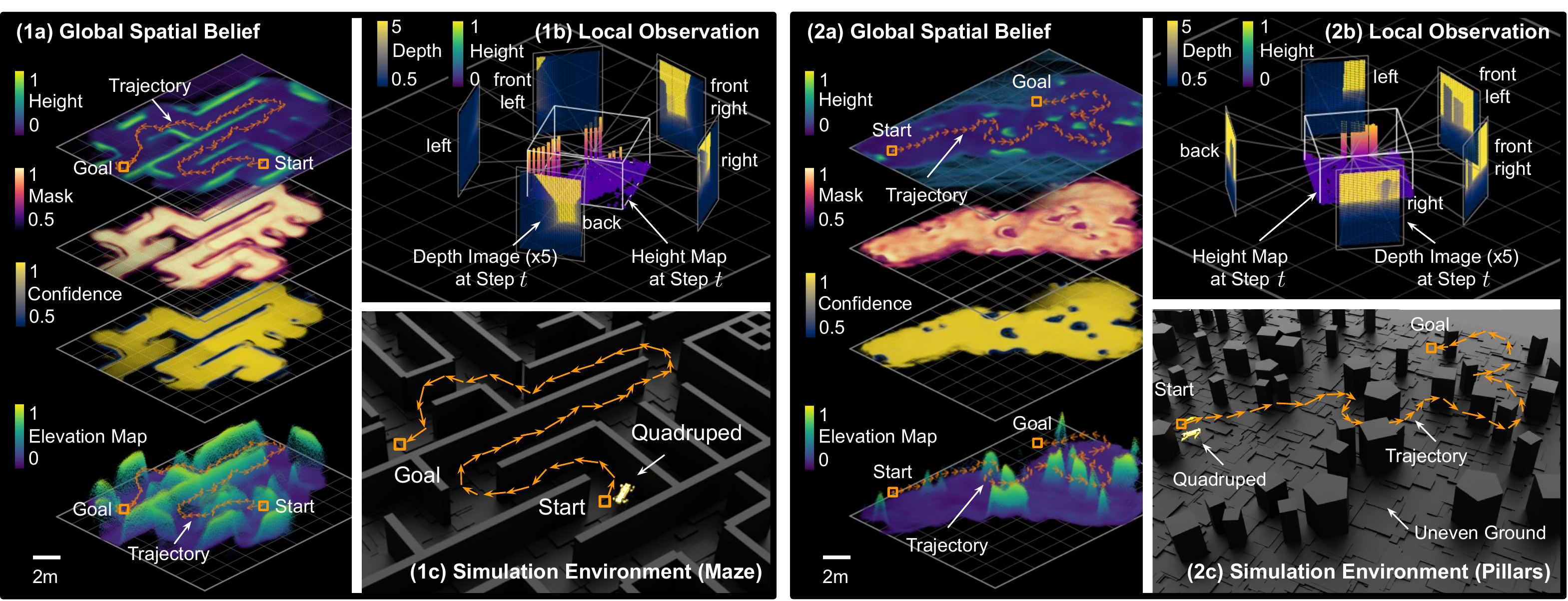}
\caption{
Visualization of learned spatial beliefs in two simulation environments: (1) maze and (2) pillars. (a) Global spatial beliefs reconstructed from predicted height, mask, and confidence maps, with robot trajectories overlaid. (b) Local observations obtained by fusing depth images from five cameras (front-left, front-right, left, right, and rear) into local height maps. (c) Simulation environments with the quadruped robot and executed trajectories overlaid.
}
    \label{fig:global_spatial_belief}
\end{figure}

\begin{table*}[t]
\centering
\caption{
Navigation results over 3 independent runs of 1000-episode simulations.
Failure modes include timeout (TO), collision (Coll.), and excessive roll (Roll).
Best SR and SPL are bolded and underlined, while lowest failure rates are underlined.
}
\label{tab:combined_results}

\vspace{1mm}

{\small (a) Model Comparison on Chained-Goal Navigation}

\vspace{1mm}

\resizebox{\textwidth}{!}{
\begin{tabular}{l c c c ccccc ccccc}
\toprule

\multirow{3}{*}{\textbf{Method}}
& \multirow{3}{*}{
\begin{tabular}{c}
\textbf{\# Params} \\
\textbf{(M)}
\end{tabular}
}
& \multirow{3}{*}{\textbf{Memory}}
& \multirow{3}{*}{\textbf{Stats}}
& \multicolumn{5}{c}{\textbf{Path Length 5$\times$[4,15] m}}
& \multicolumn{5}{c}{\textbf{Path Length 5$\times$[4,20] m}}
\\

\cmidrule(lr){5-9}
\cmidrule(lr){10-14}

&&&&
\textbf{SR $\uparrow$} &
\textbf{SPL $\uparrow$} &
\textbf{TO $\downarrow$} &
\textbf{Coll. $\downarrow$} &
\textbf{Roll $\downarrow$}
&
\textbf{SR $\uparrow$} &
\textbf{SPL $\uparrow$} &
\textbf{TO $\downarrow$} &
\textbf{Coll. $\downarrow$} &
\textbf{Roll $\downarrow$}
\\

&&&&
\textbf{(\%)} & \textbf{} & \textbf{(\%)} & \textbf{(\%)} & \textbf{(\%)}
&
\textbf{(\%)} & \textbf{} & \textbf{(\%)} & \textbf{(\%)} & \textbf{(\%)}
\\

\midrule

\multirow{2}{*}{Ours}
& \multirow{2}{*}{$5.69$}
& \multirow{2}{*}{GLST}
& Mean
& $\underline{\mathbf{79.2}}$
& $\underline{\mathbf{0.70}}$
& $\underline{10.1}$
& $\underline{9.1}$
& $1.6$
& $\underline{\mathbf{64.6}}$
& $\underline{\mathbf{0.56}}$
& $\underline{20.4}$
& $\underline{12.9}$
& $2.1$
\\

&&& Std
& $0.8$ & $0.01$ & $0.6$ & $0.9$ & $0.5$
& $1.8$ & $0.02$ & $1.1$ & $0.7$ & $0.3$
\\

\midrule

\multirow{2}{*}{SRU~\cite{yang2025spatially}}
& \multirow{2}{*}{11.8}
& \multirow{2}{*}{\cite{yang2025spatially}}
& Mean
& $64.2$ & $0.54$ & $15.8$ & $19.1$ & $\underline{0.9}$
& $50.3$ & $0.42$ & $21.2$ & $27.9$ & $\underline{0.6}$
\\

&&& Std
& $2.8$ & $0.03$ & $2.1$ & $1.5$ & $0.4$
& $3.2$ & $0.03$ & $3.0$ & $1.8$ & $0.4$
\\

\addlinespace[2.5pt]

\multirow{2}{*}{PCGrad~\cite{yu2020gradient}}
& \multirow{2}{*}{10.1}
& \multirow{2}{*}{Unified}
& Mean
& $55.3$ & $0.40$ & $18.7$ & $20.9$ & $5.1$
& $45.6$ & $0.32$ & $24.4$ & $26.5$ & $3.5$
\\

&&& Std
& $4.9$ & $0.04$ & $2.3$ & $2.2$ & $3.5$
& $3.3$ & $0.03$ & $2.9$ & $2.7$ & $2.3$
\\

\midrule

\multirow{2}{*}{Transformer}
& \multirow{2}{*}{10.0}
& \multirow{2}{*}{Unified}
& Mean
& $49.1$ & $0.33$ & $24.6$ & $24.5$ & $1.8$
& $43.7$ & $0.29$ & $23.5$ & $30.8$ & $2.0$
\\

&&& Std
& $2.3$ & $0.02$ & $2.1$ & $1.6$ & $0.6$
& $2.6$ & $0.02$ & $2.6$ & $2.0$ & $0.9$
\\

\addlinespace[2.5pt]

\multirow{2}{*}{LSTM}
& \multirow{2}{*}{10.1}
& \multirow{2}{*}{Unified}
& Mean
& $48.9$ & $0.33$ & $25.8$ & $22.4$ & $2.9$
& $39.7$ & $0.26$ & $26.3$ & $30.9$ & $3.1$
\\

&&& Std
& $2.5$ & $0.02$ & $2.4$ & $2.0$ & $0.4$
& $3.0$ & $0.02$ & $3.1$ & $2.2$ & $0.6$
\\

\bottomrule
\end{tabular}
}

\vspace{2mm}

{\small (b) Ablation Study}

\vspace{1mm}

\resizebox{\textwidth}{!}{
\begin{tabular}{c c c c c c c c c c c c c c c}
\toprule

\multicolumn{4}{c}{\textbf{Modules}} &
\multicolumn{1}{c}{} &
\multicolumn{5}{c}{\textbf{Path Length 5$\times$[4,15] m}} &
\multicolumn{5}{c}{\textbf{Path Length 5$\times$[4,20] m}} \\

\cmidrule(lr){1-4}
\cmidrule(lr){6-10}
\cmidrule(lr){11-15}

\textbf{Global} & \textbf{Local} & \textbf{Temporal} & \textbf{Rewards} & \textbf{Stats}
& \textbf{SR $\uparrow$} & \textbf{SPL $\uparrow$} & \textbf{TO $\downarrow$} & \textbf{Coll. $\downarrow$} & \textbf{Roll $\downarrow$}
& \textbf{SR $\uparrow$} & \textbf{SPL $\uparrow$} & \textbf{TO $\downarrow$} & \textbf{Coll. $\downarrow$} & \textbf{Roll $\downarrow$} \\

&&&&
& \textbf{(\%)} & & \textbf{(\%)} & \textbf{(\%)} & \textbf{(\%)}
& \textbf{(\%)} & & \textbf{(\%)} & \textbf{(\%)} & \textbf{(\%)} \\

\midrule

\checkmark & \checkmark & \checkmark & \checkmark & Mean
& $\underline{\mathbf{79.2}}$ & $\underline{\mathbf{0.70}}$ & $\underline{10.1}$ & $9.1$ & $1.6$
& $\underline{\mathbf{64.6}}$ & $\underline{\mathbf{0.56}}$ & $\underline{20.4}$ & $12.9$ & $2.1$ \\

\checkmark & \checkmark & \checkmark & \checkmark & Std
& $0.8$ & $0.01$ & $0.6$ & $0.9$ & $0.5$
& $1.8$ & $0.02$ & $1.1$ & $0.7$ & $0.3$ \\

\midrule

$-$ & \checkmark & \checkmark & \checkmark & Mean
& $74.0$ & $0.66$ & $16.9$ & $\underline{8.3}$ & $\underline{0.8}$
& $54.1$ & $0.46$ & $34.5$ & $\underline{10.2}$ & $\underline{1.2}$ \\

$-$ & \checkmark & \checkmark & \checkmark & Std
& $2.6$ & $0.02$ & $2.8$ & $0.4$ & $0.2$
& $3.5$ & $0.03$ & $3.1$ & $0.7$ & $0.1$ \\

\midrule

\checkmark & $-$ & \checkmark & \checkmark & Mean
& $55.7$ & $0.48$ & $25.3$ & $13.3$ & $5.6$
& $49.6$ & $0.41$ & $27.0$ & $16.2$ & $7.2$ \\

\checkmark & $-$ & \checkmark & \checkmark & Std
& $1.2$ & $0.01$ & $1.0$ & $0.1$ & $1.0$
& $1.6$ & $0.01$ & $1.3$ & $1.9$ & $2.7$ \\

\midrule

\checkmark & \checkmark & $-$ & \checkmark & Mean
& $60.6$ & $0.54$ & $16.4$ & $20.3$ & $2.6$
& $46.2$ & $0.44$ & $23.1$ & $27.7$ & $3.0$ \\

\checkmark & \checkmark & $-$ & \checkmark & Std
& $2.4$ & $0.02$ & $1.0$ & $1.4$ & $0.1$
& $3.0$ & $0.08$ & $2.8$ & $4.2$ & $0.4$ \\

\midrule

\checkmark & \checkmark & \checkmark & $-$ & Mean
& $70.7$ & $0.65$ & $18.7$ & $9.6$ & $1.0$
& $51.3$ & $0.46$ & $30.8$ & $15.8$ & $2.0$ \\

\checkmark & \checkmark & \checkmark & $-$ & Std
& $3.0$ & $0.03$ & $2.6$ & $0.1$ & $0.3$
& $2.0$ & $0.02$ & $0.9$ & $1.5$ & $0.5$ \\

\bottomrule
\end{tabular}
}

\end{table*}


We evaluate our model across four terrain types: maze environments, pillar obstacles, pits, and multilevel stair environments. Fig.~\ref{fig:ours_performance_breakdown} summarizes chained-goal navigation performance in the maze environment (see Appendix~\ref{app:simulation_quantitative_results} for all terrains and varied path lengths). In Fig.~\ref{fig:ours_performance_breakdown}(a), chained success rates are evaluated up to 5 sequential goals by comparing our chained-goal formulation with independent-goal execution. Success exhibits multiplicative decay as the number of goals increases. However, the GLST framework reduces the final failure rate from 39\% to 21\%.

Fig.~\ref{fig:ours_performance_breakdown}(b) suggests that single-goal success rates increase across the goal sequence. Timeout and collision dominate, while excessive roll remains rare. Timeouts primarily arise from suboptimal routing under partial observations, whereas collisions often arise from aggressive corner-cutting and perception limitations. Fig.~\ref{fig:ours_performance_breakdown}(c) evaluates navigation efficiency using SPL. SPL improves across the goal sequence despite comparable shortest-path distributions, suggesting that the efficiency gains are not simply attributable to shorter or easier navigation paths.

We further examine learned spatial beliefs produced by the global spatial module. During inference, only latent beliefs are used for navigation; decoded maps are reconstructed solely for visualization and analysis. Figure~\ref{fig:global_spatial_belief} presents examples in maze and pillar environments. The learned global belief does not preserve fine-grained geometry, but instead captures coarse traversability structures and major environmental layouts. Local height maps are recursively integrated to update the global belief representation. Despite long detours and local minima, the reconstructed beliefs preserve sufficient structure for successful navigation. In both maze and pillar environments, the robot consistently identifies traversable openings and navigates through constrained passages. 


Baseline comparisons are summarized in Table~\ref{tab:combined_results}(a). We consider two sets of baselines. First, SRU~\cite{yang2025spatially} represents a state-of-the-art method for long-horizon mapless navigation over challenging terrains, while PCGrad~\cite{yu2020gradient} and CAGrad~\cite{liu2021conflict} evaluate whether gradient-conflict mitigation can address the competing memory requirements of exploration and strategic revisitation. We report PCGrad as the stronger of the two gradient-based baselines. GLST improves SR over SRU by over 14\% absolute (23\% relative) and over PCGrad by at least 19\% absolute (41\% relative), with substantially fewer timeout and collision failures. Unlike gradient-based methods that mitigate conflicts during optimization, GLST explicitly separates global, local, and temporal representations in memory.
Second, we compare against transformer and recurrent architectures with unified memory, intentionally configured with more parameters than GLST to control for model capacity. GLST consistently achieves higher SR and SPL, supporting that its gains arise from the proposed spatio-temporal belief decomposition with rewards design rather than increased model capacity. Implementation details and separate evaluations of the two online chained-goal navigation scenarios for GLST and the baselines are provided in Appendix~\ref{app:baseline}.

The ablation study in Table~\ref{tab:combined_results}(b) evaluates the contributions of the proposed architecture and goal-conditioned rewards. Removing the global spatial module degrades performance, particularly under longer path settings, highlighting the importance of persistent global memory for revisitation and long-horizon reasoning. Removing the local spatial module causes the largest performance drop, demonstrating the importance of local traversability representations for reactive navigation. Temporal and exploration reward ablations exhibit consistent trends, further supporting the effectiveness of the proposed global-local spatio-temporal architecture.

\begin{figure}[t]
    \centering
    \includegraphics[width=1\linewidth]{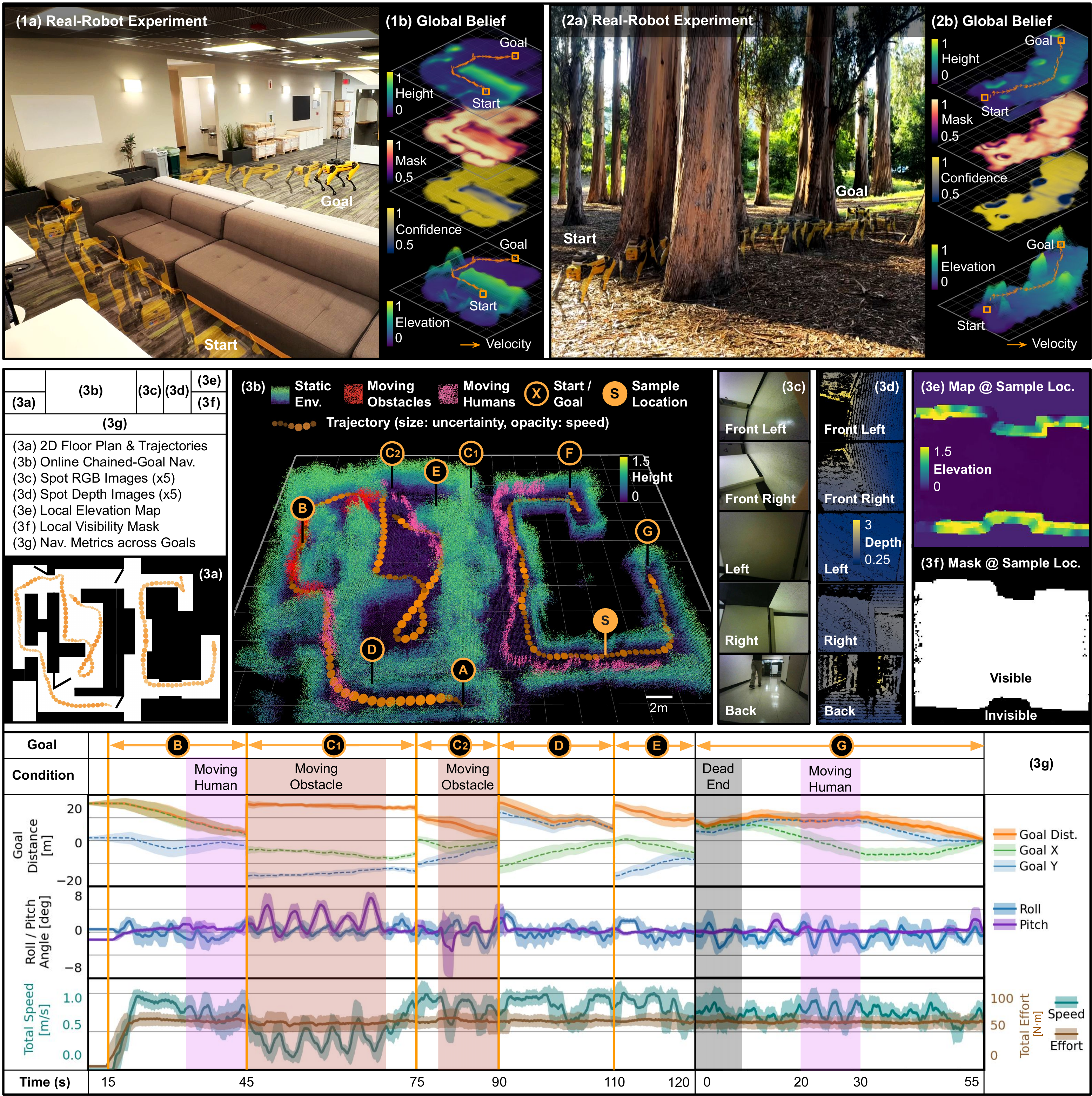}
\caption{
Real-robot navigation experiments on the Spot quadruped across diverse indoor and outdoor environments. (1) Indoor experiments. (2) Outdoor forest experiments. (a) Experimental scenes with ghost robot overlays illustrating executed trajectories. (b) Inferred global spatial beliefs reconstructed from latent representations. (3) Long-horizon online chained-goal experiments with dead ends and dynamic obstacles. Additional chained-goal indoor and outdoor experimental results are provided in Appendix~\ref{app:realworld_exp}.
}
    \label{fig:real_world_exp}
\end{figure}

\subsection{Real-World Experiments}
We conduct real-robot experiments using the Spot quadruped in indoor and outdoor environments. The policy is deployed on an onboard Jetson Thor. It receives height scans, odometry, and goal observations, and outputs velocity commands to the built-in Spot controller. The height scans are sampled from an elevation map generated from onboard depth data, while odometry is provided by the Spot SDK. Fig.~\ref{fig:real_world_exp}(1--2) presents representative indoor and outdoor experiments with vision-based body obstacle avoidance enabled, together with reconstructed global spatial beliefs inferred from latent representations. Despite relying only on local observations, the learned belief captures coarse traversability structures and preserves sufficient environmental context.

Fig.~\ref{fig:real_world_exp}(3) presents a more complex long-horizon online chained-goal navigation experiment spanning multiple rooms, with dead ends, dynamic obstacles, and body obstacle avoidance disabled. The shortest-path distance to each goal can exceed 30\,m, while moving obstacles and humans interfere with the robot during navigation. Goals are provided sequentially during execution: B, D, E, and G are newly emerging goals, while C$_1$ and C$_2$ redirect the current goal. Fig.~\ref{fig:real_world_exp}(3b) shows the executed trajectory and a representative sample location (S); the corresponding RGB and depth observations are shown in Fig.~\ref{fig:real_world_exp}(3c--3d), and the resulting local elevation map and visibility mask in Fig.~\ref{fig:real_world_exp}(3e--3f). Fig.~\ref{fig:real_world_exp}(3g) shows navigation metrics across goals. 
Goal distance is measured as the 2D Euclidean distance and may therefore decrease or increase as Spot follows a traversable path toward each goal, particularly in dead-end and maze-like environments.
Pitch exhibits the largest attitude variations during interactions with moving obstacles and humans, accompanied by fluctuations in total speed, while total effort, measured as the joint-torque norm $\|\boldsymbol{\tau}\|_2$ over the 12 leg joints, remains relatively stable.
We repeat the real-robot evaluation for 36 trials per method and compare against baselines. Our method achieves an 86.1\% success rate, outperforming the best baseline, \cite{yang2025spatially}, by 11.1 percentage points in SR and by 0.09 in SPL across the evaluated terrain layouts, while achieving comparable speed (0.61 vs.~0.62\,m/s). GLST succeeds in dead-end and chained-goal scenarios where the baselines struggle. Implementation details and additional chained-goal experiments 
across varied terrains are provided in Appendix~\ref{app:realworld_exp}.

\section{Limitations}
(a) Since GLST uses the local height map for both global and local spatial modules, its performance is sensitive to the quality of this input. In real-robot experiments, strong interactions with movable objects causing large pitch or roll deviations can introduce drift in the Spot SDK height map, potentially saturating velocity commands.
(b) The proposed goal-conditioned memory is primarily empirically motivated. Rigorous analysis of conflicting memory gradients and quantitative evaluation of their resolution remain important future directions.
(c) GLST adopts hierarchical rather than joint locomotion-navigation training, with the locomotion policy pretrained and frozen during navigation training. Consequently, some failures are difficult to attribute to navigation or locomotion limitations. Joint training and sensitivity analysis of navigation performance across different locomotion policies remain important future directions.
(d) Sim-to-real gaps remain challenging. Obstacles such as chairs and tables can be difficult to perceive reliably, while our map representation does not correctly capture 3D structures such as overhangs. A gap also persists between simulated map characteristics and those perceived on the real robot.
\section{Conclusion}
\label{sec:conclusion}
In this work, we introduced GLST, a global-local spatio-temporal belief learning framework for online chained-goal navigation under partial observability and evolving objectives. We identified conflicting memory requirements between exploration and strategic revisitation arising from ambiguous goal evolution, and addressed this challenge through explicit spatio-temporal belief decomposition together with goal-conditioned rewards. Extensive simulation experiments across diverse terrains demonstrated improved long-horizon navigation performance, while real-robot experiments on the Spot quadruped demonstrated robust long-horizon chained-goal navigation across challenging indoor and outdoor environments, including dead ends and dynamic obstacles, while reconstructing global spatial beliefs from only local observations.

\acknowledgments{
We thank the Area Chair and anonymous reviewers for their constructive feedback. This research was supported in part by the Defense Advanced Research Projects Agency (DARPA) under Agreement No. HR00112490425. The views, opinions, and/or findings expressed are those of the author(s) and should not be interpreted as representing the official views or policies of the Defense Advanced Research Projects Agency (DARPA), or the U.S. Government.
}

\clearpage
\bibliography{final}  

\appendix

\newpage

\section{Estimation of Chained-Goal Success Rate Using Monte Carlo Simulation Under the Assumption of Independent-Goal Strategy}
\label{app:chained_sr}

We estimate chained-goal success rates from the single-goal success rates reported in prior work under the assumption of an independent-goal strategy, where the policy does not retain belief memory across goals. The retaining-belief strategy is not considered in this analysis. For each method, let
\begin{align*}
\mathcal{P}=\{p_1,\dots,p_K\}
\end{align*}
denote the set of reported single-goal success rates. For a chain of length $n_g$, we generate Monte Carlo samples by independently and uniformly drawing
\begin{align*}
\tilde p_1,\dots,\tilde p_{n_g}
\sim
\mathrm{Uniform}(\mathcal{P}).
\end{align*}

For each sampled chain, the chained-goal success rate is computed as
\begin{align*}
\mathrm{SR}_{\mathrm{chain}}
=
\prod_{j=1}^{n_g}
\tilde p_j.
\end{align*}

Repeating this procedure yields an empirical distribution of chained-goal success rates for each goal index $n_g$.

\begin{figure}[h]
    \centering
    \includegraphics[width=1\linewidth]{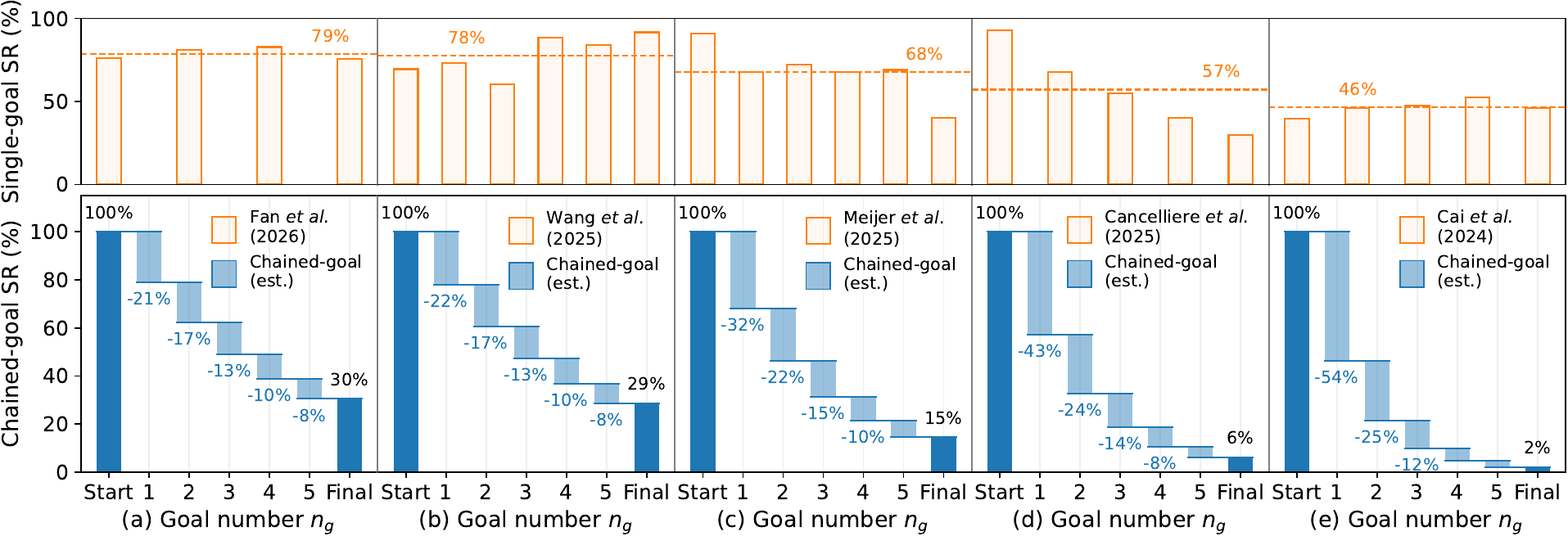}
    \caption{Chained-goal success rate (\%) as a function of goal index $n_g$, where $n_g=1,\dots,5$ denotes the $n_g$-th goal in the sequence. The lower figures show the compounded success rate from 100\% to the final success rate under the assumption of independent goal completion. The upper figures show the single-goal success rates reported in prior work, which are used to construct the chained-goal success rates. Data are adapted from (a)~\cite{yang2025spatially}, (b)~\cite{wang2025x}, (c)~\cite{meijer2025pushing}, (d)~\cite{cancelliere2025reinforcement}, and (e)~\cite{cai2024deep}.}
    \label{fig:literature_chained_goal}
\end{figure}

In Fig.~\ref{fig:literature_chained_goal}, the upper row shows the reported single-goal success rates used as the sampling pool. The lower row visualizes the estimated chained-goal success rate obtained from the Monte Carlo procedure, where each waterfall step corresponds to the average reduction caused by adding one additional goal to the sequence. This analysis estimates how single-goal performance compounds under sequential goal completion when individual goal outcomes are assumed to be independent.

The reported success rates in prior work vary substantially because the cited methods were evaluated under different terrains, navigation horizons, success and failure criteria, robotic embodiments, and sensing modalities. Therefore, this analysis is not intended as a direct comparison of performance across methods. Instead, the purpose is to illustrate the rapid success-rate degradation that can emerge in chained-goal navigation under an independent-goal strategy. In contrast, when belief memory is retained across goals, the conflicting exploration and exploitation gradients discussed in the problem formulation section may arise.

\newpage
\section{Proof of Conflicting Gradients in Online Chained-Goal Navigation}
\label{app:conflicting_gradients}

In online chained-goal navigation, goals may evolve during execution.
As a result, optimal trajectories may partially overlap with previously explored regions, requiring both continued exploration and revisitation under partial observability. The resulting chained-goal objective is
\begin{align*}
\pi_{\mathrm{nav}}^{*}
&=
\arg\max_{\pi} J(\theta)
=
\arg\max_{\pi}
\underset{\tau}{\mathbb{E}}
\left[
\sum_{n=1}^{N_g}
\sum_{t=1}^{T_n}
{r}_{n,t}
\right].
\end{align*}


where $r_{n,t}$ is the reward at step $(n,t)$. The policy gradient can be written as
\begin{align*}
\nabla_{\theta} J
=
\mathbb{E}
\left[
\sum_{n=1}^{N_g}
\sum_{t=1}^{T_n}
\nabla_{\theta}
\log
\pi_{\theta}
(a_{n,t}\mid o_{n,t})
\sum_{n=1}^{N_g}
\mathcal{R}(\tau_{n})
\right],
\end{align*}
where $\mathcal{R}(\tau_{n})$ is the cumulative return for goal $n$.

To analyze the conflicting behaviors induced by exploration and revisitation, the objective can be decomposed into exploration and revisitation components:
\begin{align*}
J(\theta)
=
J^{+}(\theta)
+
J^{-}(\theta).
\end{align*}

The exploration reward encourages trajectories to remain dissimilar from previously visited regions through state-distribution divergence. We model each visited position using a localized kernel occupancy distribution, while trajectory dissimilarity is measured using the Jensen--Shannon (JS) divergence:
\begin{align*}
\mathcal{R}^{+}(\tau_n)
&=
\lambda_{+}
\sum_{t=1}^{T_n}
\sum_{(m,k)\prec(n,t)}
D_{\mathrm{JS}}
\left(
p_{n,t}
\|
p_{m,k}
\right),
\end{align*}
where
$
p_{n,t}(x)
=
\kappa(x,x_{n,t})
$
represents a localized state occupancy distribution around position $x_{n,t}$.

Conversely, the revisitation reward encourages trajectories to align with previously useful paths by minimizing state-distribution divergence:
\begin{align*}
\mathcal{R}^{-}(\tau_n)
&=
-\lambda_{-}
\sum_{t=1}^{T_n}
\min_{1\le j\le L_n}
D_{\mathrm{JS}}
\left(
p_{n,t}
\|
p^{*}_{n,j}
\right),
\end{align*}
where
$
p^{*}_{n,j}(x)
=
\kappa(x,x^{*}_{n,j}).
$ Here, $x^{*}_{n,j}$ denotes the $j$-th state along the reference trajectory.

We consider the common case where the optimal trajectory toward a new goal partially overlaps with regions visited during previous goals. Assume that the optimal reference trajectory for goal $n$ overlaps with a previously visited state distribution. That is, there exist $(m,k)\prec(n,t)$ and $j$ such that
\begin{align*}
d_{\mathrm{TV}}
\left(
p^{*}_{n,j},
p_{m,k}
\right)
\le
\epsilon,
\end{align*}
where $d_{\mathrm{TV}}(\cdot,\cdot)$ denotes the total variation distance, measuring the similarity between previously visited and reference trajectory state distributions.

Define
\begin{align*}
\Delta_{n,t,j,m,k}
&=
D_{\mathrm{JS}}
\left(
p_{n,t}
\|
p^{*}_{n,j}
\right)
-
D_{\mathrm{JS}}
\left(
p_{n,t}
\|
p_{m,k}
\right).
\end{align*}

By continuity of the Jensen--Shannon divergence with respect to its arguments,
\begin{align*}
\left|
\Delta_{n,t,j,m,k}
\right|
\le
\omega(\epsilon),
\end{align*}
where $\omega(\epsilon)\to 0$ as $\epsilon\to 0$.

Therefore, the divergence between the current state distribution and the reference trajectory becomes approximately equal to the divergence from the previously visited distribution:
\begin{align*}
D_{\mathrm{JS}}
\left(
p_{n,t}
\|
p^{*}_{n,j}
\right)
=
D_{\mathrm{JS}}
\left(
p_{n,t}
\|
p_{m,k}
\right)
+
O(\omega(\epsilon)).
\end{align*}

Thus, around the overlapping region, denote $\mathcal{R}^{-}$ as the local revisitation reward: 
\begin{align*}
\mathcal{R}^{-}
&\approx
-\lambda_{-}
D_{\mathrm{JS}}
\left(
p_{n,t}
\|
p_{m,k}
\right),
\end{align*}
while the local exploration reward, $\mathcal{R}^{+}$, contains
\begin{align*}
\mathcal{R}^{+}
&\approx
+\lambda_{+}
D_{\mathrm{JS}}
\left(
p_{n,t}
\|
p_{m,k}
\right).
\end{align*}

Denote $\mathcal{J}^+$, $\mathcal{J}^-$ as the local objectives induced by local rewards $\mathcal{R}^+$ and $\mathcal{R}^-$, respectively. 
\begin{align*}
\nabla \mathcal{J}^{-}
&=
\mathbb{E}
\left[
\sum_{n=1}^{N_g}
\sum_{t=1}^{T_n}
\nabla_{\theta}
\log
\pi_{\theta}
(a_{n,t}\mid o_{n,t})
\mathcal{R}^-
\right] \\
&\approx -\lambda_{-}
\mathbb{E}
\left[
\sum_{n=1}^{N_g}
\sum_{t=1}^{T_n}
\nabla_{\theta}
\log
\pi_{\theta}
(a_{n,t}\mid o_{n,t})
D_{\mathrm{JS}}
\left(
p_{n,t}
\|
p_{m,k}
\right)
\right] 
\end{align*}
Similarly, we can have
\begin{align*}
\nabla \mathcal{J}^{+}
&\approx +\lambda_{+}
\mathbb{E}
\left[
\sum_{n=1}^{N_g}
\sum_{t=1}^{T_n}
\nabla_{\theta}
\log
\pi_{\theta}
(a_{n,t}\mid o_{n,t})
D_{\mathrm{JS}}
\left(
p_{n,t}
\|
p_{m,k}
\right)
\right] 
\end{align*}






Therefore,
\begin{align*}
\left\langle
\nabla \mathcal{J}^{+},
\nabla \mathcal{J}^{-}
\right\rangle
&\approx
-\lambda_{+}\lambda_{-}
\left\|
\mathbb{E}
\left[
\sum_{n=1}^{N_g}
\sum_{t=1}^{T_n}
\nabla_{\theta}
\log
\pi_{\theta}
(a_{n,t}\mid o_{n,t})
D_{\mathrm{JS}}
\left(
p_{n,t}
\|
p_{m,k}
\right)
\right]
\right\|
^2 
<0,
\end{align*}
whenever $\lambda_{+}>0$, $\lambda_{-}>0$, and
\begin{align*} 
\left\|
\mathbb{E}
\left[
\sum_{n=1}^{N_g}
\sum_{t=1}^{T_n}
\nabla_{\theta}
\log
\pi_{\theta}
(a_{n,t}\mid o_{n,t})
D_{\mathrm{JS}}
\left(
p_{n,t}
\|
p_{m,k}
\right)
\right]
\right\|
\neq 0.
\end{align*}

Hence, in the local regions where the optimal path overlaps with previously visited states, exploration and revisitation induce opposing gradient directions:
\begin{align*}
\boxed{
\left\langle
\nabla \mathcal{J}^{+},
\nabla \mathcal{J}^{-}
\right\rangle
<
0
}.
\end{align*}

This shows that a shared memory representation can receive conflicting learning signals: exploration discourages revisitation, while exploitation encourages revisiting previously useful states.

\newpage
\section{Global-Local Spatio-Temporal Belief Learning (GLST)}
\label{app:architecture}

\begin{figure}[h]
    \centering
    \includegraphics[width=1\linewidth]{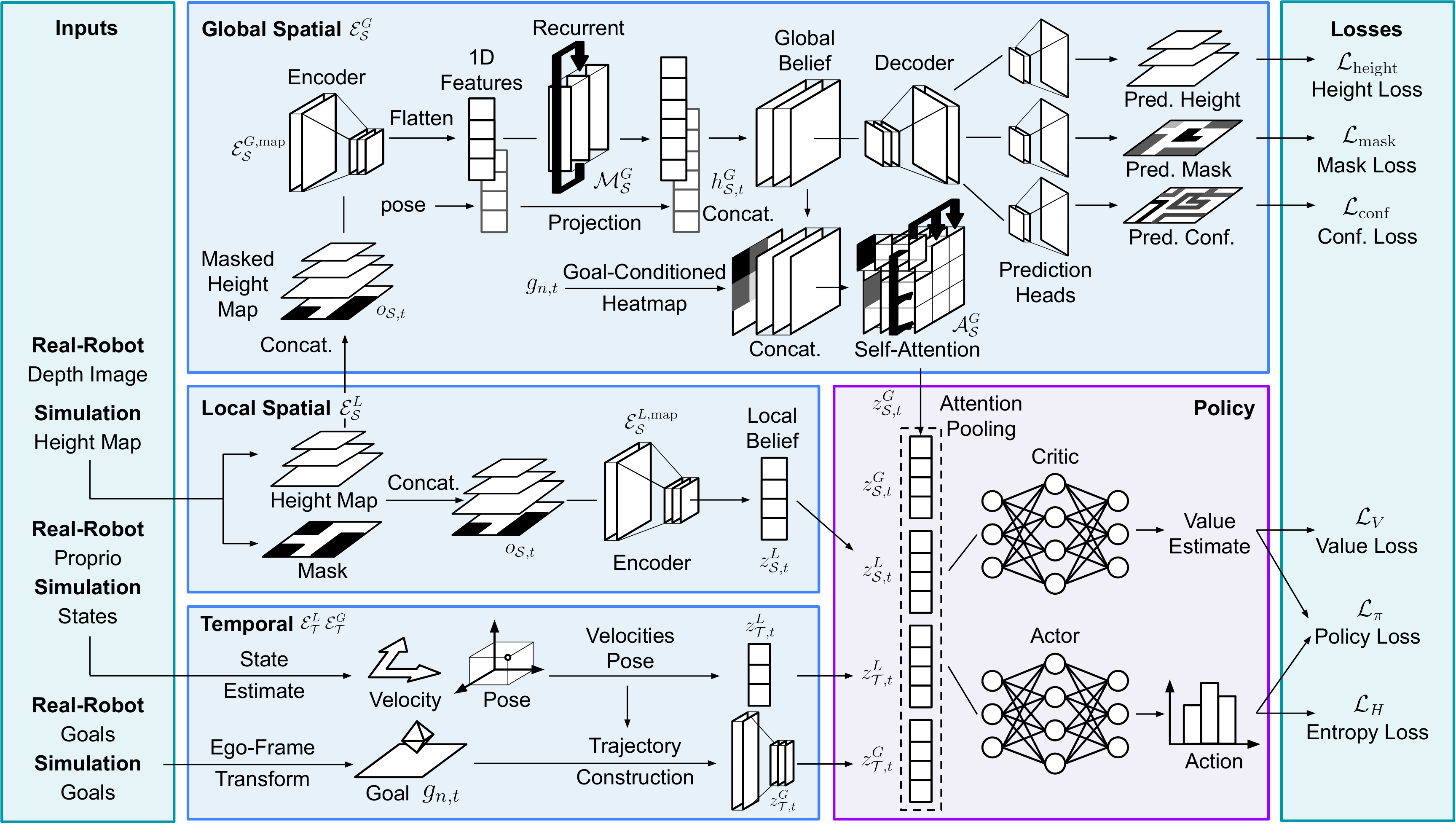}
\caption{
Detailed architecture of the proposed GLST framework. Observations are decomposed into global-local spatial and temporal belief representations. Global beliefs are recursively accumulated and refined through goal-conditioned attention, while fused belief representations are processed by a shared actor-critic policy. Auxiliary prediction heads supervise spatial belief learning through height, mask, and confidence reconstruction.
}
    \label{fig:architecture}
\end{figure}

GLST decomposes latent beliefs into complementary global-local spatial and long-short temporal representations to address the distinct memory requirements arising in long-horizon chained-goal navigation. Rather than relying on a single latent memory to implicitly learn all navigation behaviors, GLST explicitly separates nearby reactive reasoning from persistent environment memory, and short-horizon motion dynamics from long-horizon trajectory history. This decomposition allows the policy to simultaneously preserve detailed local traversability information while maintaining accumulated knowledge useful for exploration, revisitation, and trajectory reuse.

\paragraph{Global and Local Spatial Belief Learning}
The spatial module consists of local and global spatial belief encoders, $\mathcal{E}_{\mathcal{S}}^L$ and $\mathcal{E}_{\mathcal{S}}^G$, respectively. We further elaborate on global spatial belief learning in this appendix. The global spatial memory consists of two components: recurrent memory and projection-based memory.
The recurrent memory, implemented using LSTM~\cite{hochreiter1997long}, captures long-range dependencies and retains information from past observations, while the differentiable map builder accumulates spatially structured environment information over extended navigation horizons. The resulting recurrent and projection-based global memories are concatenated as $h_{\mathcal S,t}^{G}=[\tilde h_{\mathcal S,t}^{G},\hat h_{\mathcal S,t}^{G}]$ and processed by a goal-conditioned attention encoder to produce the global spatial belief representation $z_{\mathcal S,t}^{G}=\mathcal A_{\mathcal S}^{G}(h_{\mathcal S,t}^{G},g_t)$. Specifically, the accumulated latent global map is compressed into spatial tokens and augmented with positional embeddings. A self-attention encoder models interactions across spatial regions to capture long-range dependencies and global spatial relationships. Finally, a learned attention pooling operation conditioned on the navigation goal selectively aggregates spatial features into a compact global belief representation. Goal conditioning allows the same accumulated environment memory to support different navigation behaviors under changing objectives by emphasizing spatial regions relevant to the current goal.

To preserve meaningful spatial organization during representation learning, the global spatial branch is pretrained using supervised auxiliary objectives. The complete spatial memory construction process is summarized in Algorithm~\ref{alg:global_aux_pretraining}. Specifically, the fused global latent belief $h_{\mathcal S,t}^{G}$ is decoded through prediction heads for terrain reconstruction, visibility estimation, and confidence estimation:
$\hat H_t^G=\mathcal D_h(h_{\mathcal S,t}^{G})$,
$\hat M_t^G=\mathcal D_m(h_{\mathcal S,t}^{G})$,
and
$\hat C_t^G=\mathcal D_c(h_{\mathcal S,t}^{G})$.

The complete auxiliary objective combines terrain reconstruction, visibility prediction, and confidence estimation:
\begin{align}
\mathcal L
=
\mathcal L_{\mathrm{height}}
+
\lambda_m\mathcal L_{\mathrm{mask}}
+
\lambda_c\mathcal L_{\mathrm{conf}} .
\end{align}
where $\lambda_m$ and $\lambda_c$ weight the visibility and confidence objectives.

The terrain reconstruction objective applies supervision only over visible regions while combining squared and absolute reconstruction penalties:
\begin{align}
\ell^H_{b,t,i,j}
=
(1-\alpha_h)
(\hat H^G_{b,t,i,j}-H^G_{b,t,i,j})^2
+
\alpha_h
|\hat H^G_{b,t,i,j}-H^G_{b,t,i,j}|,
\end{align}
\begin{align}
\mathcal L_{\mathrm{height}}
=
\frac{1}{BT}
\sum_{b=1}^{B}
\sum_{t=1}^{T}
\frac{
\sum_{i,j}
M^G_{b,t,i,j}\ell^H_{b,t,i,j}
}{
\sum_{i,j}M^G_{b,t,i,j}
},
\end{align}
where $B$ and $T$ denote the batch size and sequence length, $(i,j)$ denotes spatial grid coordinates, $\hat H^G$ and $H^G$ denote predicted and ground-truth global height maps, $M^G$ denotes the ground-truth visibility mask, and $\alpha_h$ balances squared and absolute reconstruction penalties. The combined reconstruction penalty exploits the smooth optimization behavior of squared error while using absolute error to better preserve sharp terrain discontinuities and reduce over-smoothed predictions near obstacle boundaries and height transitions.

Visibility prediction uses binary cross entropy:
\begin{align}
\mathcal L_{\mathrm{mask}}
=
\frac{1}{BT}
\sum_{b=1}^{B}
\sum_{t=1}^{T}
\sum_{i,j}
\left[
-
M^G_{b,t,i,j}\log \hat P^G_{b,t,i,j}
-
(1-M^G_{b,t,i,j})
\log(1-\hat P^G_{b,t,i,j})
\right],
\end{align}
where $\hat M^G$ denotes predicted visibility logits, and $\hat P^G=\sigma(\hat M^G)$ denotes predicted visibility probability. Binary cross entropy is used because visibility prediction is formulated as a per-grid-location binary classification problem, where each map location indicates whether the corresponding region has been observed.

Confidence targets are generated from reconstruction quality:
\begin{align}
C^G_{b,t,i,j}
=
\exp
\left(
-\alpha_c
\left(
\hat H^G_{b,t,i,j}\hat P^G_{b,t,i,j}
-
H^G_{b,t,i,j}M^G_{b,t,i,j}
\right)^2
\right),
\end{align}
\begin{align}
\mathcal L_{\mathrm{conf}}
=
\frac{1}{BT}
\sum_{b=1}^{B}
\sum_{t=1}^{T}
\sum_{i,j}
(\hat C^G_{b,t,i,j}-C^G_{b,t,i,j})^2,
\end{align}
where $\alpha_c$ controls the sensitivity of confidence targets to reconstruction error, $\hat C^G$ denotes the predicted confidence map and $C^G$ denotes the confidence target. Confidence targets are generated from reconstruction quality so that regions with accurate and consistent predictions receive higher confidence, encouraging the model to learn uncertainty estimates that reflect the reliability of accumulated spatial memories.

These auxiliary objectives encourage the global spatial memory to preserve geometric consistency, visibility structure, and uncertainty estimates while learning latent representations suitable for downstream policy optimization.

\begin{algorithm}[h]
\caption{Global Spatial Encoder Pretraining}
\label{alg:global_aux_pretraining}
\small
\begin{algorithmic}[1]
\Require Episode dataset $\mathbb D=\{(o_{1:T},p_{1:T},H^G_{1:T},M^G_{1:T})\}$
\Require Local height map encoder $\mathcal{E}_{\mathcal{S}}^{G,\mathrm{map}}$,
recurrent memory module $\tilde{\mathcal{M}}_{\mathcal{S}}^G$,
projection-based memory module $\hat{\mathcal{M}}_{\mathcal{S}}^G$,
decoder heads $\mathcal D_h, \mathcal D_m, \mathcal D_c$

\For{each training epoch}
    \For{each episode batch}
        \State Initialize global latent memory $\mathcal B$ using initial poses $p_1$
        \State Initialize recurrent memory state $(h_0,c_0)$

        \For{$t=1,\dots,T$}
            \State Encode local observation:
            $u^L_{\mathcal{S},t}
            =\mathcal{E}_{\mathcal{S}}^{G,\mathrm{map}}(o_{\mathcal{S},t})$

            \State Update recurrent memory:
            $\tilde{h}^{G}_{\mathcal{S},t}
            =\tilde{\mathcal{M}}_{\mathcal{S}}^G
            \left(\tilde{h}^{G}_{\mathcal{S},t-1},
            u^{L}_{\mathcal{S},t},p_t\right)$

            \State Project local latent features into the global frame:
            $\hat{h}_{\mathcal{S},t}^G
            =\hat{\mathcal{M}}_{\mathcal{S}}^G
            \left(\hat{h}^{G}_{\mathcal{S},t-1},
            u^{L}_{\mathcal{S},t},p_t\right)$

            \State Fuse recurrent and projection-based global memories:
            $h^{G}_{\mathcal{S},t}
            =[\tilde h_{\mathcal{S},t}^G,
            \hat h_{\mathcal{S},t}^G]$

            \State Decode height, mask, and confidence:
            $\hat H_t^G=\mathcal D_h(h^{G}_{\mathcal{S},t}),\;
            \hat M_t^G=\mathcal D_m(h^{G}_{\mathcal{S},t}),\;
            \hat C_t^G=\mathcal D_c(h^{G}_{\mathcal{S},t})$
        \EndFor

        \State Compute supervised loss:
        \State $\mathcal L=
        \mathcal L_{\mathrm{height}}(\hat H^G,H^G,M^G)
        +\lambda_m\mathcal L_{\mathrm{mask}}(\hat M^G,M^G)
        +\lambda_c\mathcal L_{\mathrm{conf}}(\hat C^G,C^G)$

        \State Update $\mathcal{E}_{\mathcal{S}}^{G,\mathrm{map}},
        \tilde{\mathcal{M}}_{\mathcal{S}}^G,
        \mathcal D_h,\mathcal D_m,\mathcal D_c$
        by backpropagation
    \EndFor
\EndFor
\end{algorithmic}
\end{algorithm}

\paragraph{Global and Local Temporal Belief Learning}

The temporal branch similarly separates short-horizon and long-horizon reasoning through local and global temporal belief encoders,
$\mathcal E_{\mathcal T}^{L}$
and
$\mathcal E_{\mathcal T}^{G}$,
respectively. The local temporal belief
$z_{\mathcal T,t}^{L}$
captures short-term motion dynamics using instantaneous robot states, including robot velocities and goal observations. This representation primarily supports reactive motion adaptation and immediate control decisions.

In contrast, the global temporal belief
$z_{\mathcal T,t}^{G}$
preserves long-horizon trajectory information accumulated across chained goals. To explicitly support both exploration and trajectory reuse, GLST maintains three trajectory memories with distinct roles. The current trajectory memory stores short-horizon trajectory observations associated with the active goal and is reset whenever a new chained subgoal occurs. This memory captures recent motion behavior relevant to immediate navigation decisions. The accumulated trajectory memory continuously records trajectory observations throughout chained-goal execution without resetting across subgoal transitions. It therefore acts as a persistent internal history preserving long-horizon exploration information across multiple goals. The historical snapshot memory stores a copy of the accumulated trajectory memory immediately before a goal transition. Consequently, it preserves trajectories associated with previously completed subgoals while excluding information generated during the current goal. At goal transitions, the accumulated trajectory memory is first snapshotted into the historical memory, after which the current trajectory memory is reinitialized while continuous accumulation proceeds. This separation prevents recent trajectories from overwriting persistent navigation context while allowing the policy to reason jointly over current motion history and previously explored trajectories. For relatively short trajectory histories and reliable state estimation, raw trajectory observations may already provide sufficient temporal information. However, under noisy localization or substantially longer temporal horizons, separate 1D CNN trajectory encoders may further improve robustness by extracting higher-level temporal patterns and compressing long-horizon trajectory dependencies. Goal transitions can be detected using either a predefined rule or a learned inference model. In the rule-based approach, transitions are identified based on changes in the relative robot-to-goal distance. In the inference-based approach, only the current goal coordinates are observed, without explicit goal IDs or transition/success signals. An MLP infers goal transitions from the history of relative robot-to-goal observations.

\paragraph{PPO Implementation Details} 

For navigation policy training, we generate procedurally varied terrains consisting of four categories with distinct navigation challenges: maze terrains with constrained corridors and dead ends, stair terrains with elevation transitions and height discontinuities, pit terrains with traversability gaps and constrained pathways, and pillar-obstacle terrains with densely distributed obstacles requiring precise local maneuvering. These terrain types expose the policy to diverse traversability patterns and improve robustness across varying geometric structures and long-horizon navigation scenarios.

For each environment, we construct a collision-aware traversability graph. Valid terrain points are sampled and filtered by wall clearance, height constraints, and collision checks. Nearby valid points are connected into a graph, which is used to construct paths and calculate shortest-path lengths. Training goals are sampled from the traversability graph rather than directly based on Euclidean distance. Start-goal pairs are selected according to the curriculum path-length range. During chained-goal training, subsequent goals are sampled online through feasible graph-based transitions, ensuring reachability while generating long-horizon navigation tasks with trajectory reuse and exploration requirements. Sampled nodes with graph degree lower than three are removed during graph construction to improve the feasibility of subsequent online goal sampling. The training and evaluation framework additionally supports configurable constraints on subsequent goal sampling. For example, candidate goals can be filtered based on the angle between the previous and subsequent goal-direction vectors, allowing chained-goal scenarios to favor confined-space navigation, extended exploration, or unconstrained random goal transitions. To further reduce the sim-to-real gap for long-range dead-end escape, we additionally randomize maze geometry by varying wall thickness up to $1\,\mathrm{m}$, which improves dead-end navigation. Further improvements could potentially be achieved by augmenting the current location-triggered position history with temporally triggered history, providing denser historical information during extended dead-end escape.


During training, navigation policies are optimized using short chained-goal sequences consisting of two goals with path lengths between 4~m and 10~m, while evaluation considers path lengths of up to 20~m with up to five goals. Despite this limited training setting, the learned policy generalizes effectively to substantially longer chained-goal trajectories, demonstrating the ability of GLST to preserve and reuse long-horizon spatial and temporal information beyond the training horizon.

Policy optimization is performed using proximal policy optimization (PPO)~\cite{schulman2017proximal} with generalized advantage estimation (GAE)~\cite{schulman2015high} through the RSL-RL~\cite{schwarke2025rsl} backend in Isaac Lab. The policy uses a non-recurrent actor-critic architecture; temporal information is provided through the GLST observation representation rather than through an additional PPO recurrent module. The actor and critic share the same encoded observation structure but use separate MLP heads with hidden dimensions $[512,256,128]$ and ELU activations. The policy input is the GLST belief representation
$z_t=\{z^{L}_{\mathcal S,t},z^{G}_{\mathcal S,t},z^{L}_{\mathcal T,t},z^{G}_{\mathcal T,t}\}$,
which combines local spatial, global spatial, local temporal, and global temporal beliefs before being passed to the actor and critic networks. In implementation, these beliefs are instantiated using encoded local height-map features, goal-conditioned latent global-map features, instantaneous robot/goal observations, and chained trajectory-history observations.

Training uses rollouts of 48 steps per environment for 1500 learning iterations. PPO uses clipped policy updates with clipping parameter $0.2$, clipped value loss, GAE with $\gamma=0.99$ and $\lambda=0.95$, value loss coefficient $0.5$, entropy coefficient $0.005$, 5 learning epochs, 4 minibatches per update, Adam optimization with initial learning rate $10^{-3}$, adaptive KL scheduling with target KL $0.01$, and gradient clipping with maximum norm $1.0$. Empirical observation normalization is disabled. No additional PPO auxiliary loss is used during policy optimization.


\newpage
\section{Rewards and Curriculum}
\label{app:rewards_and_curriculum}

This appendix provides additional details on the reward formulation used during training. Goal-progress and exploration rewards are presented in the main text; here, we provide further discussion of these rewards and their alternative formulations, along with the remaining reward terms and curriculum design.

\paragraph{Progress Reward.} One limitation of the progress reward design is that rewards are assigned whenever a new closest distance is achieved, even if the robot has entered a local optimum. Consequently, the robot may first collect nearby progress rewards even when local observations already suggest that the current direction is suboptimal. Time penalties mitigate this behavior by discouraging inefficient trajectories. However, under partial observability, collecting immediately achievable rewards remains preferable because future opportunities are uncertain and discounted over time.

\paragraph{Exploration Reward.}
In the main text, we present a continuous formulation of the exploration reward, where previously visited positions contribute according to a Gaussian kernel. This provides a smoothly varying penalty with respect to spatial distance. Alternatively, we can use a discrete neighborhood-based formulation, which penalizes revisitation only within a predefined neighborhood and avoids contributions from sufficiently distant states. The discrete formulation is
\begin{align}
\mathcal{R}_e(t)
=
-
\lambda_e
\left(
\min(n_{n,t},n_{\max})
\right)^{\gamma}
\sum_{\mathbf{x}_i\in\mathcal B_n}
\mathbf{1}
\left\{
\|\mathbf{x}_t-\mathbf{x}_i\|_2 < r_e
\right\},
\end{align}
where $\lambda_e$ controls the penalty magnitude, $n_{n,t}$ denotes the number of nearby visits around the current position $\mathbf{x}_t$, $n_{\max}$ bounds the revisit scaling, $\gamma$ controls its growth, and $r_e$ denotes the neighborhood radius. The indicator function $\mathbf{1}\{\cdot\}$ equals one when a stored position $\mathbf{x}_i$ lies within $r_e$ of the current position and zero otherwise. The cap $n_{\max}$ can be omitted when an unbounded increase in the revisit penalty is desired. Similar to the continuous formulation, the visitation buffer $\mathcal B_n$ is reset at each goal transition, such that revisitation is penalized within the current goal segment without interfering with exploration under subsequent goals.

\paragraph{Collision penalty.}

To discourage collisions and excessive impacts, we penalize contact-force magnitude acting on the robot torso. Let $f_t\ge0$ denote the contact force magnitude and let $\tau$ be a tolerance threshold below which incidental contacts are ignored. The penalty can be generalized as
\begin{align}
\mathcal R_{\mathrm{contact}}(t)
=
-\lambda_c
\max(f_t-\tau,0)^p,
\end{align}
where $\lambda_c>0$ controls penalty magnitude and $p\ge1$ controls growth rate. In implementation, this penalty can either be applied directly or approximated using a piecewise form with multiple force thresholds and penalty weights to more strongly penalize larger impacts.

\paragraph{Time penalty.}

To encourage efficient navigation and discourage idling or unnecessary detours, we apply a constant per-step penalty:
\begin{align}
\mathcal R_{\mathrm{time}}(t)
=
-\lambda_t,
\end{align}
where $\lambda_t\geq 0$ is a weighting factor. This simple penalty discourages standing still and oscillatory behavior while complementing progress rewards.

\paragraph{Task reward.}

To provide sparse task-level supervision without introducing dense shaping signals, we adopt the time-gated stochastic reward from prior work~\cite{yang2025spatially, he2024agile}. Let $x_t$ denote the robot position, $g_{n,t}$ denote the current goal position, and $T_{\max}$ denote the episode horizon. A reward signal is activated either within the final $T_r$ seconds of the episode or with random probability $\delta_{\mathrm{check}}$:
\begin{align}
\mathcal R_{\mathrm{task}}(t)
=
\frac{
\mathbf{1}
\left\{
t>T_{\max}-T_r
\;\text{or}\;
u_t<\delta_{\mathrm{check}}
\right\}
}{
1+
\|x_t-g_{n,t}\|_2/\sigma
},
\qquad
u_t\sim
\mathrm{Unif}(0,1),
\end{align}
where $T_r$ is the terminal reward window, $\delta_{\mathrm{check}}$ is the random checking probability, and $\sigma$ controls spatial sensitivity. This reward remains predominantly sparse while providing occasional task-level guidance, and is complemented by dense shaping rewards.

\paragraph{Lateral movement penalty.}
To discourage unnecessary sideways motion, we penalize lateral velocity in the robot body frame. Let $v_{y,t}^{b}$ denote the body-frame lateral velocity. The penalty is
\begin{align}
\mathcal R_{\mathrm{lat}}(t)
=
-\lambda_{\mathrm{lat}}
\operatorname{clip}
\left(
(v_{y,t}^{b})^2,
0,
1
\right),
\end{align}
where $\lambda_{\mathrm{lat}}>0$ controls the penalty magnitude.

\paragraph{Action-rate regularization.}
To encourage smooth commands, we penalize deviations from a momentum-filtered action. Let $a_t$ denote the current navigation action and define
\begin{align}
a_t^{m}
=
\eta a_{t-1}^{m}
+
(1-\eta)a_t,
\end{align}
where $\eta\in[0,1)$ is the momentum coefficient. The action-rate penalty is
\begin{align}
\mathcal R_{\mathrm{act}}(t)
=
-\lambda_{\mathrm{act}}
\left(
\beta_1
\|a_t-a_t^{m}\|_1
+
\beta_2
\|\ddot q_t\|_1
\right),
\end{align}
where $\lambda_{\mathrm{act}}\geq 0$ controls the penalty magnitude, $\beta_1$ and $\beta_2$ weight the action deviation and joint acceleration terms, respectively, and $\ddot q_t$ denotes joint acceleration.

\paragraph{Velocity penalty.}
To avoid overly aggressive motion, we penalize planar speed:
\begin{align}
\mathcal R_{\mathrm{vel}}(t)
=
-\lambda_{\mathrm{vel}}
\|v_{xy,t}^{w}\|_2,
\end{align}
where $\lambda_{\mathrm{vel}}\geq 0$ controls the penalty magnitude and $v_{xy,t}^{w}$ is the robot planar velocity in the world frame.

\paragraph{Success bonus.}
Finally, we include a sparse success bonus when the goal-reaching termination condition is satisfied:
\begin{align}
\mathcal R_{\mathrm{succ}}(t)
=
\lambda_{\mathrm{succ}}
\mathbf{1}_{\mathrm{goal}}(t),
\end{align}
where $\lambda_{\mathrm{succ}}\geq 0$ controls the bonus magnitude, and $\mathbf{1}_{\mathrm{goal}}(t)=1$ if the goal is reached at timestep $t$ and $0$ otherwise.

Among all reward terms, the progress rewards, exploration reward, collision penalty, and time penalty have the largest impact on navigation performance and success rate. Reward weights are task-dependent and should be selected according to the desired behavior, such as rapid goal reaching, obstacle avoidance, or efficient exploration. Additional terms, including lateral movement penalties and action-rate regularization, primarily improve locomotion quality by encouraging smoother and more stable gait behaviors.

\paragraph{Curriculum.}
We employ a curriculum strategy to stabilize training for long-horizon chained-goal navigation. The curriculum progressively increases navigation difficulty through terrain-level progression and goal-distance expansion. Specifically, goal path lengths are gradually increased from shorter ranges to longer ranges during training, while terrain levels are advanced according to policy performance. We also randomize the initial robot heading to improve robustness to diverse starting orientations. In addition, the vertical goal-progress reward is adjusted during training to improve learning on multilevel terrain. This curriculum reduces early-stage exploration failures and enables the policy to gradually adapt to longer horizons and more complex environments.


\newpage

\section{Detailed Chained-Goal Evaluation Results}
\label{app:simulation_quantitative_results}

Table~\ref{tab:chained_goal_results} summarizes chained-goal navigation performance across different terrains and path-length settings. Overall success rates decrease as path length and terrain complexity increase, while per-goal SR and SPL generally improve across chained stages, suggesting that trajectory reuse and accumulated spatial context improve navigation efficiency. 
The path-length statistics reported here compute standard deviation across random seeds, whereas the path-length statistics in Figure~\ref{fig:ours_performance_breakdown} report standard deviation across all episodes. Subsequent goals may have fewer episodes when no valid chained goal is available in the sampled navigation graph.

\begin{table*}[h]
\centering
\caption{Chained-goal navigation statistics across terrains and path-length settings. Results are aggregated over 3 runs of 1000-episode simulations and reported as mean $\pm$ std across runs.}
\label{tab:chained_goal_results}
\vspace{2mm}
\resizebox{\textwidth}{!}{
\begin{tabular}{l l c c c c c c c c}
\toprule
\textbf{Terrain}
& \textbf{Goal}
& \textbf{Path length}
& \textbf{\# Episodes}
& \textbf{Speed}
& \textbf{SR (\%) $\uparrow$}
& \textbf{SPL $\uparrow$}
& \multicolumn{3}{c}{\textbf{Failure modes (\%)}} \\
\cmidrule(lr){8-10}
& & \textbf{(m)} & & \textbf{(m/s)} & & &
\textbf{Timeout $\downarrow$} & \textbf{Collision $\downarrow$} & \textbf{Roll $\downarrow$} \\
\midrule

\multicolumn{10}{c}{\textbf{Path Length: $5\times[4,15]$ m}} \\
\midrule

\multirow{6}{*}{Maze}
& Overall & $46.4 \pm 0.7$ & $1000$ & $1.1$ & $79.2 \pm 0.9$ & $0.70 \pm 0.01$ & $10.1 \pm 0.7$ & $9.1 \pm 1.1$ & $1.6 \pm 0.5$ \\
& G1 & $9.3 \pm 0.3$ & $1000$ & $1.0$ & $90.6 \pm 0.3$ & $0.84 \pm 0.01$ & $5.0 \pm 0.9$ & $4.0 \pm 0.8$ & $0.4 \pm 0.2$ \\
& G2 & $9.2 \pm 0.3$ & $906 \pm 3$ & $1.1$ & $94.4 \pm 0.5$ & $0.85 \pm 0.01$ & $3.2 \pm 0.2$ & $2.0 \pm 0.2$ & $0.4 \pm 0.2$ \\
& G3 & $9.4 \pm 0.0$ & $856 \pm 5$ & $1.1$ & $96.5 \pm 0.7$ & $0.86 \pm 0.01$ & $1.9 \pm 0.4$ & $1.2 \pm 0.4$ & $0.2 \pm 0.1$ \\
& G4 & $9.5 \pm 0.1$ & $825 \pm 4$ & $1.1$ & $97.5 \pm 0.6$ & $0.88 \pm 0.01$ & $0.8 \pm 0.2$ & $1.4 \pm 0.3$ & $0.4 \pm 0.2$ \\
& G5 & $9.2 \pm 0.5$ & $805 \pm 4$ & $1.1$ & $98.4 \pm 0.3$ & $0.89 \pm 0.01$ & $0.0 \pm 0.1$ & $1.4 \pm 0.2$ & $0.2 \pm 0.1$ \\

\midrule
\multirow{6}{*}{Pillars}
& Overall & $55.4 \pm 0.1$ & $1000$ & $1.1$ & $95.9 \pm 1.2$ & $0.94 \pm 0.01$ & $1.2 \pm 0.3$ & $1.7 \pm 0.5$ & $1.0 \pm 0.3$ \\
& G1 & $10.8 \pm 0.1$ & $1000$ & $1.0$ & $97.5 \pm 0.9$ & $0.95 \pm 0.01$ & $0.9 \pm 0.3$ & $0.9 \pm 0.5$ & $0.3 \pm 0.1$ \\
& G2 & $11.0 \pm 0.1$ & $973 \pm 8$ & $1.1$ & $99.0 \pm 0.3$ & $0.96 \pm 0.01$ & $0.2 \pm 0.1$ & $0.3 \pm 0.1$ & $0.2 \pm 0.1$ \\
& G3 & $11.0 \pm 0.1$ & $963 \pm 8$ & $1.1$ & $99.3 \pm 0.2$ & $0.96 \pm 0.01$ & $0.0 \pm 0.1$ & $0.2 \pm 0.1$ & $0.4 \pm 0.2$ \\
& G4 & $11.1 \pm 0.1$ & $958 \pm 6$ & $1.1$ & $99.6 \pm 0.1$ & $0.96 \pm 0.01$ & $0.1 \pm 0.1$ & $0.2 \pm 0.1$ & $0.1 \pm 0.1$ \\
& G5 & $11.2 \pm 0.1$ & $955 \pm 6$ & $1.1$ & $99.7 \pm 0.1$ & $0.96 \pm 0.01$ & $0.0 \pm 0.0$ & $0.1 \pm 0.1$ & $0.1 \pm 0.1$ \\

\midrule
\multirow{6}{*}{Stairs}
& Overall & $56.3 \pm 0.3$ & $1000$ & $1.1$ & $80.2 \pm 1.3$ & $0.76 \pm 0.01$ & $1.4 \pm 0.5$ & $12.0 \pm 0.4$ & $6.4 \pm 0.8$ \\
& G1 & $11.2 \pm 0.1$ & $1000$ & $1.0$ & $94.2 \pm 0.5$ & $0.88 \pm 0.00$ & $0.8 \pm 0.2$ & $4.1 \pm 0.5$ & $1.0 \pm 0.1$ \\
& G2 & $11.2 \pm 0.1$ & $900 \pm 10$ & $1.1$ & $95.6 \pm 0.9$ & $0.91 \pm 0.01$ & $0.4 \pm 0.2$ & $2.1 \pm 0.4$ & $1.9 \pm 2.4$ \\
& G3 & $11.4 \pm 0.2$ & $863 \pm 10$ & $1.1$ & $95.9 \pm 0.5$ & $0.90 \pm 0.01$ & $0.1 \pm 0.1$ & $2.4 \pm 0.4$ & $1.3 \pm 0.5$ \\
& G4 & $11.2 \pm 0.0$ & $825 \pm 9$ & $1.1$ & $95.6 \pm 0.9$ & $0.90 \pm 0.01$ & $0.0 \pm 0.1$ & $2.7 \pm 0.3$ & $1.8 \pm 0.5$ \\
& G5 & $11.4 \pm 0.1$ & $802 \pm 12$ & $1.1$ & $97.2 \pm 0.4$ & $0.92 \pm 0.00$ & $0.0 \pm 0.0$ & $1.4 \pm 0.3$ & $1.0 \pm 0.2$ \\

\midrule
\multirow{6}{*}{Pits}
& Overall & $54.5 \pm 0.3$ & $1000$ & $1.1$ & $89.4 \pm 1.4$ & $0.87 \pm 0.01$ & $0.8 \pm 0.3$ & $4.5 \pm 0.8$ & $5.3 \pm 0.6$ \\
& G1 & $10.9 \pm 0.2$ & $1000$ & $1.0$ & $95.2 \pm 1.0$ & $0.92 \pm 0.01$ & $0.7 \pm 0.3$ & $1.8 \pm 0.6$ & $3.0 \pm 0.6$ \\
& G2 & $10.9 \pm 0.1$ & $952 \pm 12$ & $1.1$ & $98.2 \pm 0.8$ & $0.95 \pm 0.01$ & $0.0 \pm 0.0$ & $0.7 \pm 0.2$ & $1.4 \pm 0.3$ \\
& G3 & $11.0 \pm 0.2$ & $904 \pm 20$ & $1.1$ & $97.8 \pm 0.5$ & $0.95 \pm 0.01$ & $0.1 \pm 0.1$ & $1.0 \pm 0.2$ & $1.0 \pm 0.2$ \\
& G4 & $10.9 \pm 0.2$ & $890 \pm 18$ & $1.1$ & $98.3 \pm 0.5$ & $0.95 \pm 0.01$ & $0.0 \pm 0.0$ & $0.7 \pm 0.2$ & $0.8 \pm 0.4$ \\
& G5 & $11.0 \pm 0.1$ & $887 \pm 13$ & $1.1$ & $98.4 \pm 0.7$ & $0.95 \pm 0.01$ & $0.0 \pm 0.0$ & $0.6 \pm 0.3$ & $0.8 \pm 0.6$ \\

\midrule
\midrule
\multicolumn{10}{c}{\textbf{Path Length: $5\times[4,20]$ m}} \\
\midrule

\multirow{6}{*}{Maze}
& Overall & $65.8 \pm 2.2$ & $1000$ & $1.1$ & $64.6 \pm 2.3$ & $0.56 \pm 0.02$ & $20.4 \pm 1.1$ & $12.9 \pm 0.7$ & $2.1 \pm 0.4$ \\
& G1 & $12.5 \pm 0.1$ & $1000$ & $1.0$ & $85.0 \pm 1.7$ & $0.78 \pm 0.01$ & $9.2 \pm 1.5$ & $5.3 \pm 1.0$ & $0.5 \pm 0.1$ \\
& G2 & $13.5 \pm 0.3$ & $766 \pm 17$ & $1.1$ & $90.1 \pm 0.9$ & $0.80 \pm 0.00$ & $6.0 \pm 1.0$ & $3.2 \pm 0.4$ & $0.7 \pm 0.3$ \\
& G3 & $13.3 \pm 0.5$ & $706 \pm 19$ & $1.1$ & $92.1 \pm 0.9$ & $0.82 \pm 0.01$ & $3.5 \pm 0.4$ & $3.0 \pm 0.5$ & $0.4 \pm 0.1$ \\
& G4 & $13.4 \pm 0.7$ & $667 \pm 18$ & $1.1$ & $94.5 \pm 0.6$ & $0.84 \pm 0.01$ & $1.8 \pm 0.4$ & $2.4 \pm 0.4$ & $0.7 \pm 0.1$ \\
& G5 & $13.6 \pm 0.5$ & $646 \pm 18$ & $1.1$ & $96.8 \pm 0.3$ & $0.85 \pm 0.00$ & $0.5 \pm 0.0$ & $2.2 \pm 0.3$ & $0.3 \pm 0.1$ \\

\midrule
\multirow{6}{*}{Pillars}
& Overall & $73.5 \pm 0.4$ & $1000$ & $1.1$ & $93.2 \pm 1.4$ & $0.90 \pm 0.03$ & $1.4 \pm 0.5$ & $4.1 \pm 1.3$ & $0.9 \pm 0.3$ \\
& G1 & $14.6 \pm 0.1$ & $1000$ & $1.0$ & $97.0 \pm 0.8$ & $0.93 \pm 0.01$ & $0.7 \pm 0.1$ & $1.7 \pm 0.4$ & $0.3 \pm 0.2$ \\
& G2 & $14.8 \pm 0.1$ & $967 \pm 7$ & $1.1$ & $98.4 \pm 0.3$ & $0.94 \pm 0.01$ & $0.5 \pm 0.3$ & $0.7 \pm 0.2$ & $0.1 \pm 0.1$ \\
& G3 & $14.6 \pm 0.1$ & $948 \pm 7$ & $1.1$ & $98.7 \pm 0.3$ & $0.94 \pm 0.01$ & $0.1 \pm 0.1$ & $0.9 \pm 0.1$ & $0.2 \pm 0.1$ \\
& G4 & $14.7 \pm 0.2$ & $939 \pm 5$ & $1.1$ & $99.1 \pm 0.2$ & $0.94 \pm 0.01$ & $0.0 \pm 0.1$ & $0.7 \pm 0.2$ & $0.1 \pm 0.1$ \\
& G5 & $14.8 \pm 0.1$ & $930 \pm 6$ & $1.1$ & $99.0 \pm 0.4$ & $0.94 \pm 0.01$ & $0.0 \pm 0.0$ & $0.8 \pm 0.3$ & $0.1 \pm 0.1$ \\

\midrule
\multirow{6}{*}{Stairs}
& Overall & $72.5 \pm 0.2$ & $1000$ & $1.1$ & $67.2 \pm 1.5$ & $0.63 \pm 0.02$ & $3.9 \pm 2.0$ & $19.6 \pm 1.3$ & $9.3 \pm 0.8$ \\
& G1 & $14.5 \pm 0.0$ & $1000$ & $1.0$ & $88.8 \pm 1.4$ & $0.83 \pm 0.01$ & $2.9 \pm 1.0$ & $7.0 \pm 1.0$ & $1.3 \pm 0.3$ \\
& G2 & $14.5 \pm 0.2$ & $843 \pm 11$ & $1.1$ & $91.7 \pm 1.1$ & $0.86 \pm 0.02$ & $0.6 \pm 0.6$ & $4.6 \pm 1.1$ & $3.0 \pm 0.8$ \\
& G3 & $14.6 \pm 0.3$ & $755 \pm 14$ & $1.1$ & $92.8 \pm 1.1$ & $0.87 \pm 0.02$ & $0.3 \pm 0.3$ & $4.8 \pm 0.8$ & $2.0 \pm 0.3$ \\
& G4 & $14.6 \pm 0.2$ & $710 \pm 14$ & $1.1$ & $94.0 \pm 0.7$ & $0.88 \pm 0.00$ & $0.3 \pm 0.6$ & $4.0 \pm 0.8$ & $2.4 \pm 0.4$ \\
& G5 & $14.7 \pm 0.2$ & $672 \pm 12$ & $1.1$ & $94.7 \pm 0.2$ & $0.89 \pm 0.00$ & $0.0 \pm 0.0$ & $3.2 \pm 0.9$ & $2.3 \pm 0.8$ \\

\midrule
\multirow{6}{*}{Pits}
& Overall & $71.4 \pm 0.4$ & $1000$ & $1.1$ & $84.0 \pm 1.4$ & $0.81 \pm 0.01$ & $1.2 \pm 0.2$ & $6.7 \pm 0.4$ & $8.0 \pm 1.4$ \\
& G1 & $14.4 \pm 0.2$ & $1000$ & $1.0$ & $94.0 \pm 0.7$ & $0.90 \pm 0.01$ & $1.0 \pm 0.1$ & $2.0 \pm 0.4$ & $3.0 \pm 0.6$ \\
& G2 & $14.2 \pm 0.2$ & $917 \pm 18$ & $1.1$ & $96.5 \pm 0.2$ & $0.93 \pm 0.01$ & $0.2 \pm 0.1$ & $1.7 \pm 0.3$ & $1.5 \pm 0.2$ \\
& G3 & $14.3 \pm 0.1$ & $883 \pm 16$ & $1.1$ & $97.4 \pm 0.7$ & $0.93 \pm 0.01$ & $0.0 \pm 0.0$ & $1.3 \pm 0.4$ & $1.2 \pm 0.3$ \\
& G4 & $14.2 \pm 0.1$ & $863 \pm 12$ & $1.1$ & $97.7 \pm 0.2$ & $0.94 \pm 0.01$ & $0.0 \pm 0.0$ & $0.9 \pm 0.1$ & $1.1 \pm 0.1$ \\
& G5 & $14.4 \pm 0.1$ & $840 \pm 12$ & $1.1$ & $97.4 \pm 0.3$ & $0.93 \pm 0.01$ & $0.0 \pm 0.0$ & $1.0 \pm 0.3$ & $1.3 \pm 0.2$ \\

\bottomrule
\end{tabular}
}
\end{table*}

\section{Baseline Selection and Implementation}
\label{app:baseline}

We consider two sets of baselines. The first set includes prior methods for long-horizon mapless navigation in challenging terrains and methods that mitigate gradient conflicts during training. The second set includes alternative architectures with unified memory representations.

In the first set of baselines, SRU~\cite{yang2025spatially} represents a state-of-the-art method with spatially enhanced memory designed for long-horizon mapless navigation over challenging terrains, while PointNav~\cite{partsey2022mapping} represents a strong method for zero-shot sim-to-real transfer that explicitly handles localization through learned odometry. We report SRU as the stronger of the two navigation-specific baselines. PCGrad~\cite{yu2020gradient} and CAGrad~\cite{liu2021conflict} are representative methods for mitigating conflicting gradients during training. They are used to evaluate whether gradient-conflict mitigation alone can address the competing memory requirements of exploration and strategic revisitation. We report PCGrad as the stronger of the two gradient-conflict-mitigation baselines. Since these methods are well detailed in the original papers, we do not further elaborate on their implementations.

\begin{table*}[!b]
\centering
\caption{
Navigation results over 1,000-rollout simulations ($\times 10$ seeds).
Failure modes include timeout (TO), collision (Coll.), and excessive roll (Roll).
Best SR and SPL are bolded and underlined, while lowest failure rates are underlined.
}
\label{tab:separate_evaluation}

\vspace{1mm}

{\small (a) New-Goal Emergence}

\vspace{1mm}

\resizebox{\textwidth}{!}{
\begin{tabular}{l c c c ccccc ccccc}
\toprule

\multirow{3}{*}{\textbf{Method}}
& \multirow{3}{*}{
\begin{tabular}{c}
\textbf{\# Params} \\
\textbf{(M)}
\end{tabular}
}
& \multirow{3}{*}{\textbf{Memory}}
& \multirow{3}{*}{\textbf{Stats}}
& \multicolumn{5}{c}{\textbf{Path Length 5$\times$[4,15] m}}
& \multicolumn{5}{c}{\textbf{Path Length 5$\times$[4,20] m}}
\\

\cmidrule(lr){5-9}
\cmidrule(lr){10-14}

&&&&
\textbf{SR $\uparrow$} &
\textbf{SPL $\uparrow$} &
\textbf{TO $\downarrow$} &
\textbf{Coll. $\downarrow$} &
\textbf{Roll $\downarrow$}
&
\textbf{SR $\uparrow$} &
\textbf{SPL $\uparrow$} &
\textbf{TO $\downarrow$} &
\textbf{Coll. $\downarrow$} &
\textbf{Roll $\downarrow$}
\\

&&&&
\textbf{(\%)} & \textbf{} & \textbf{(\%)} & \textbf{(\%)} & \textbf{(\%)}
&
\textbf{(\%)} & \textbf{} & \textbf{(\%)} & \textbf{(\%)} & \textbf{(\%)}
\\

\midrule

\multirow{2}{*}{Ours}
& \multirow{2}{*}{$5.7$}
& \multirow{2}{*}{GLST}
& Mean
& $\underline{\mathbf{79.8}}$ & $\underline{\mathbf{0.77}}$
& $\underline{7.9}$ & $\underline{10.7}$ & $1.6$
& $\underline{\mathbf{64.2}}$ & $\underline{\mathbf{0.61}}$
& $\underline{12.1}$ & $\underline{19.7}$ & $4.0$
\\

&&& Std
& $4.8$ & $0.05$ & $2.4$ & $2.5$ & $0.9$
& $5.4$ & $0.05$ & $2.6$ & $2.7$ & $1.0$
\\

\midrule

\multirow{2}{*}{SRU~\cite{yang2025spatially}}
& \multirow{2}{*}{11.8}
& \multirow{2}{*}{\cite{yang2025spatially}}
& Mean
& $64.7$ & $0.54$ & $15.0$ & $18.9$ & $\underline{1.4}$
& $50.7$ & $0.41$ & $22.0$ & $26.7$ & $\underline{0.6}$
\\

&&& Std
& $2.9$ & $0.03$ & $2.2$ & $1.4$ & $0.4$
& $3.3$ & $0.03$ & $2.7$ & $1.9$ & $0.3$
\\

\addlinespace[2.5pt]

\multirow{2}{*}{PCGrad~\cite{yu2020gradient}}
& \multirow{2}{*}{10.1}
& \multirow{2}{*}{Unified}
& Mean
& $55.7$ & $0.40$ & $18.1$ & $21.0$ & $5.3$
& $45.9$ & $0.32$ & $24.2$ & $26.5$ & $3.5$
\\

&&& Std
& $4.1$ & $0.03$ & $2.8$ & $3.1$ & $3.4$
& $2.9$ & $0.02$ & $3.5$ & $3.2$ & $2.1$
\\

\midrule

\multirow{2}{*}{Transformer}
& \multirow{2}{*}{10.0}
& \multirow{2}{*}{Unified}
& Mean
& $49.8$ & $0.32$ & $25.4$ & $23.1$ & $1.7$
& $44.4$ & $0.29$ & $22.5$ & $31.7$ & $1.4$
\\

&&& Std
& $1.9$ & $0.01$ & $2.1$ & $1.2$ & $0.5$
& $2.7$ & $0.02$ & $2.4$ & $2.1$ & $0.8$
\\

\addlinespace[2.5pt]

\multirow{2}{*}{LSTM}
& \multirow{2}{*}{10.1}
& \multirow{2}{*}{Unified}
& Mean
& $48.7$ & $0.32$ & $25.5$ & $22.5$ & $3.3$
& $39.9$ & $0.26$ & $26.0$ & $30.9$ & $3.2$
\\

&&& Std
& $2.5$ & $0.02$ & $2.1$ & $1.5$ & $0.4$
& $2.7$ & $0.02$ & $2.9$ & $2.2$ & $0.6$
\\

\bottomrule
\end{tabular}
}

\vspace{2mm}

{\small (b) Existing-Goal Redirection}

\vspace{1mm}

\resizebox{\textwidth}{!}{
\begin{tabular}{l c c c ccccc ccccc}
\toprule

\multirow{3}{*}{\textbf{Method}}
& \multirow{3}{*}{
\begin{tabular}{c}
\textbf{\# Params} \\
\textbf{(M)}
\end{tabular}
}
& \multirow{3}{*}{\textbf{Memory}}
& \multirow{3}{*}{\textbf{Stats}}
& \multicolumn{5}{c}{\textbf{Path Length 5$\times$[4,15] m}}
& \multicolumn{5}{c}{\textbf{Path Length 5$\times$[4,20] m}}
\\

\cmidrule(lr){5-9}
\cmidrule(lr){10-14}

&&&&
\textbf{SR $\uparrow$} &
\textbf{SPL $\uparrow$} &
\textbf{TO $\downarrow$} &
\textbf{Coll. $\downarrow$} &
\textbf{Roll $\downarrow$}
&
\textbf{SR $\uparrow$} &
\textbf{SPL $\uparrow$} &
\textbf{TO $\downarrow$} &
\textbf{Coll. $\downarrow$} &
\textbf{Roll $\downarrow$}
\\

&&&&
\textbf{(\%)} & \textbf{} & \textbf{(\%)} & \textbf{(\%)} & \textbf{(\%)}
&
\textbf{(\%)} & \textbf{} & \textbf{(\%)} & \textbf{(\%)} & \textbf{(\%)}
\\

\midrule

\multirow{2}{*}{Ours}
& \multirow{2}{*}{$5.7$}
& \multirow{2}{*}{GLST}
& Mean
& $\underline{\mathbf{81.4}}$ & $\underline{\mathbf{0.79}}$
& $\underline{4.9}$ & $\underline{11.6}$ & $2.1$
& $\underline{\mathbf{65.8}}$ & $\underline{\mathbf{0.63}}$
& $\underline{12.9}$ & $\underline{17.2}$ & $4.1$
\\

&&& Std
& $4.6$ & $0.05$ & $2.3$ & $2.1$ & $0.9$
& $5.2$ & $0.05$ & $2.7$ & $2.4$ & $1.1$
\\

\midrule

\multirow{2}{*}{SRU~\cite{yang2025spatially}}
& \multirow{2}{*}{11.8}
& \multirow{2}{*}{\cite{yang2025spatially}}
& Mean
& $64.4$ & $0.53$ & $16.1$ & $18.8$ & $\underline{0.7}$
& $50.4$ & $0.42$ & $20.1$ & $28.8$ & $\underline{0.7}$
\\

&&& Std
& $2.4$ & $0.02$ & $2.1$ & $1.3$ & $0.3$
& $3.3$ & $0.03$ & $3.0$ & $1.8$ & $0.4$
\\

\addlinespace[2.5pt]

\multirow{2}{*}{PCGrad~\cite{yu2020gradient}}
& \multirow{2}{*}{10.1}
& \multirow{2}{*}{Unified}
& Mean
& $56.1$ & $0.39$ & $18.8$ & $20.1$ & $5.0$
& $46.0$ & $0.32$ & $24.3$ & $26.1$ & $3.7$
\\

&&& Std
& $6.1$ & $0.04$ & $1.5$ & $1.4$ & $3.8$
& $3.4$ & $0.03$ & $2.4$ & $1.9$ & $2.7$
\\

\midrule

\multirow{2}{*}{Transformer}
& \multirow{2}{*}{10.0}
& \multirow{2}{*}{Unified}
& Mean
& $50.8$ & $0.33$ & $22.5$ & $24.8$ & $1.9$
& $44.1$ & $0.28$ & $24.1$ & $29.3$ & $2.5$
\\

&&& Std
& $2.5$ & $0.02$ & $2.2$ & $1.8$ & $0.6$
& $2.7$ & $0.02$ & $2.5$ & $2.0$ & $0.9$
\\

\addlinespace[2.5pt]

\multirow{2}{*}{LSTM}
& \multirow{2}{*}{10.1}
& \multirow{2}{*}{Unified}
& Mean
& $50.4$ & $0.33$ & $25.4$ & $21.9$ & $2.3$
& $40.4$ & $0.26$ & $26.0$ & $30.6$ & $3.0$
\\

&&& Std
& $2.3$ & $0.01$ & $2.8$ & $2.3$ & $0.4$
& $3.5$ & $0.02$ & $3.1$ & $2.4$ & $0.5$
\\

\bottomrule
\end{tabular}
}

\end{table*}

In the second set of baselines, we compare against transformer and recurrent architectures with unified memory, and provide additional implementation details to ensure fair benchmarking. All baselines are trained using the same PPO framework, training curriculum, and optimization settings as GLST. The baseline models are designed with parameter counts comparable to or exceeding that of GLST, allowing performance differences to be attributed to architectural design rather than model capacity. The baseline policy receives a unified observation representation consisting of local terrain perception, proprioceptive measurements, goal information, and navigation history. Specifically, the observations include an occlusion-aware local height scan, base linear and angular velocities, relative joint positions and velocities, a robot-centric goal command, past robot positions, recent action history, and a sinusoidal encoding of the current chained-goal subepisode. The goal command consists of the relative 3D goal position and yaw, where the position components are transformed using a signed logarithmic mapping, $\operatorname{sign}(x)\log(1+|x|)$~\cite{webber2013bi}, to improve sensitivity to distant goals while preserving local resolution. Past positions record previously visited locations at fixed distance intervals and are expressed in the current robot frame, providing a compact representation of traversal history. The action history stores recently executed velocity commands together with their associated relative displacements, while the subepisode encoding provides the current goal index within the chained-goal sequence. Unlike GLST, which explicitly separates local and global spatial-temporal memories, all observations in the baseline models are processed through a single unified memory representation.



\newpage
\section{Additional Real-World Experiments}
\label{app:realworld_exp}

\begin{figure}[h]
    \centering
    \includegraphics[width=1\linewidth]{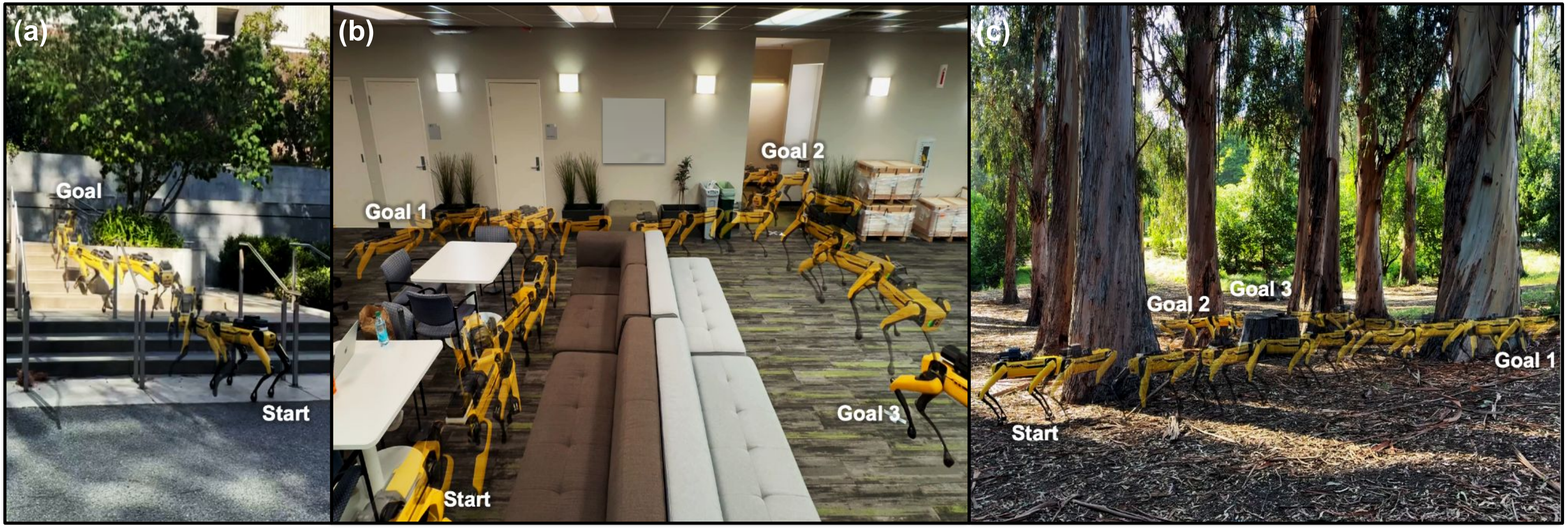}
    \caption{Additional real-robot experiments: (a) single-goal navigation on stairs, (b) indoor navigation with three chained goals, and (c) forest navigation with three chained goals. Images are contrast-enhanced for visualization.}
    \label{fig:additional_exp}
\end{figure}

We provide additional real-robot experiments in this appendix, as shown in Fig.~\ref{fig:additional_exp}, together with implementation details of the real-robot system and the measures adopted to mitigate the simulation-to-real gap.

The real-robot evaluation framework performs policy inference within a ROS2 C++ node running at 10~Hz, using elevation map data from the Spot LocalGrid at 5~Hz and system odometry at 50~Hz. To enable low-latency responses to newly observed obstacles and provide fine-grained control, the duration and magnitude of the generated velocity commands were carefully tuned before being sent to the Spot Driver, which executes the low-level locomotion controller. Perceptive collision avoidance is enabled or disabled depending on the experimental setting, as specified in the corresponding experiments. Additionally, to handle missing data within the perception pipeline, invalid values in the height scan are filled using a BST-based nearest-neighbor search, providing complete height-scan inputs to the policy. This implementation is consistent with the simulation pipeline.

Multiple discrepancies between simulation and real-world deployment contribute to performance degradation during real-robot operation. We identify three primary sources of simulation-to-real (Sim2Real) gaps.

\textbf{Height-Map Representation Constraints.}
Height-map representations introduce limitations for complex obstacle geometries. Overhanging objects may produce false-positive obstacles when sufficient clearance exists for the robot to pass underneath, while thin structures, such as chair and table legs, may be missed, resulting in false-negative obstacles. These limitations can lead to inaccurate traversability estimates in real-world environments.

\textbf{Sensor Noise and System Latency.}
Differences in sensor characteristics, ROS communication latency, and odometry noise introduce distribution shifts relative to the training environment. While our policy demonstrates robustness to moderate noise, achieving simulation-level performance in real-world deployment will likely require more extensive domain randomization to better capture these sources of uncertainty during training.

\textbf{Unmodeled Object Interactions.}
Dynamic interactions with movable objects are not modeled in simulation. The policy is trained under the assumption that obstacles remain static, and therefore selects actions without accounting for object displacement or other contact-induced changes to the environment. Interactions with movable objects can consequently produce outcomes not encountered during training, occasionally leading to suboptimal behavior or failure. Future work will investigate mitigating this Sim2Real gap through policy distillation and large-scale real-world adaptation strategies similar to those proposed by Rudin \emph{et al.}~\cite{rudin2025parkour}.

\end{document}